\documentclass{article}
\usepackage[preprint]{neurips_2025}

\usepackage{neurips_2025}

\usepackage{textcomp}
\usepackage[utf8]{inputenc} 
\usepackage[T1]{fontenc}    
\usepackage{hyperref}       
\usepackage{url}            
\usepackage{booktabs}       
\usepackage{amsfonts}       
\usepackage{nicefrac}       
\usepackage{microtype}      
\usepackage{xcolor}         
\usepackage{graphicx} 
\usepackage{url}
\usepackage{multirow}
\usepackage{amsfonts}       
\usepackage{amssymb}        
\usepackage{pifont}         
 \usepackage{amsmath} 
\newcommand{\xmark}{\ding{55}}  

\title{MoTime: A Dataset Suite for Multimodal Time Series Forecasting}

\author{%
  Xin Zhou
    \\
  Monash University\\
  Wellington Rd, Melbourne, Australia \\
  \texttt{xin.zhou@monash.edu} \\
  \And
  Weiqing Wang \\
  Monash University \\
  Wellington Rd, Melbourne, Australia \\
  \texttt{teresa.wang@monash.edu} \\
  \AND
  Francisco J. Baldán \\
  University of Málaga \\
  s/n, 29071 Málaga, Spain \\
  \texttt{fjbaldan@uma.es} \\
  \And
  Wray Buntine \\
  VinUniversity \\
  Gia Lam District, Hanoi, Vietnam \\
  \texttt{wray.b@vinuni.edu.vn} \\
  \And
  Christoph Bergmeir \\
  University of Granada \\
  s/n, 18071 Granada, Spain \\
  \texttt{bergmeir@ugr.es} \\
}

\begin{document}

\maketitle

\begin{abstract}
  While multimodal data sources are increasingly available from real-world forecasting, most existing research remains on unimodal time series. In this work, we present \texttt{MoTime}, a suite of multimodal time series forecasting datasets that pair temporal signals with external modalities such as text, metadata, and images. Covering diverse domains, \texttt{MoTime} supports structured evaluation of modality utility under two scenarios: 1) the common forecasting task where varying-length history is available, and 2) cold-start forecasting, where no historical data is available. Experiments show that external modalities can improve forecasting performance in both scenarios, with particularly strong benefits for short series in some datasets, though the impact varies depending on data characteristics.
  By making datasets and findings publicly available, we aim to support more comprehensive and realistic benchmarks in future multimodal time series forecasting research.
\end{abstract}

\section{Introduction}
Time series forecasting is an essential task across domains such as finance~\cite{bamford2023multi,sawhney2020multimodal,dong2024fnspid,djia2025}, energy~\cite{zhou2021informer,xia2024multi,wagner2020ptbxl}, retail~\cite{papadopoulos2022multimodal,yuan2022community,chang2021fashion,skenderi2024well}, and public health~\cite{li2023frozen,bai2023dreamdiffusion,johnson2016mimic,johnson2023mimiciv}. 
Recent advances have enabled researchers to scale time series corpora~\cite{makridakis2000m3,makridakis2018m4,makridakis2022m5,dua2017uci,godahewa2021monash,wang2024moirai,zhang2024moment} and train foundation models~\cite{rasul2024lagllama,das2024decoder,liang2024foundation,shi2025timemoe,liu2024moiraimoe,woo2024unified} directly on them. However, a key question remains: 
whether scaling up unimodal time series data suffices to handle the diverse real-world forecasting tasks?

\begin{figure}
    \centering
    \includegraphics[width=0.8\linewidth]{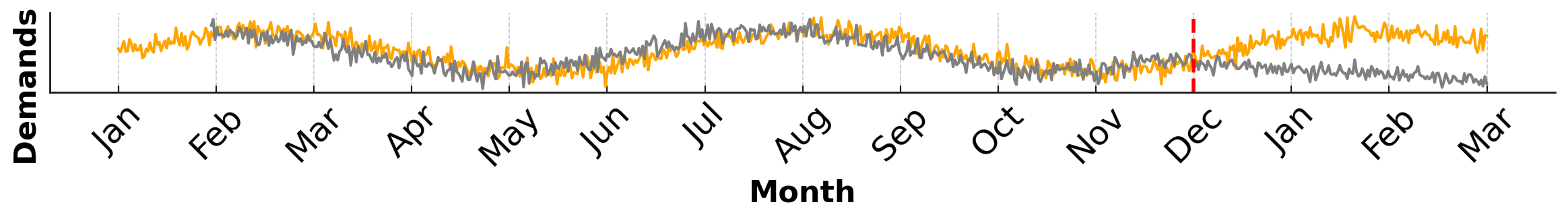}
    \caption[Monthly demand comparison]{Monthly demand of a seasonal product, portable fan (orange curve), and a trend-sensitive product, blind box (blue curve), from Jan 2023 to Mar 2024. While both series follow similar patterns in 2023, they diverge in December (red dashed line) when the blind box demand drops, highlighting the limitations of unimodal forecasting without contextual information.}
    \label{fig:motivation}
\end{figure}

We argue that the answer is not always. While large-scale time series models excel at learning recurring patterns such as seasonality or trends, they may not always be effective when the identity of the series is not directly observable from the time series alone.
Consider the example in Figure~\ref{fig:motivation}, where the demands of two items, a portable fan and a blind box toy, appear nearly identical throughout 2023. 
A unimodal time series model could generalize their behavior, treating them as governed by a similar seasonality. Yet, in early 2024, their demand diverged: the portable fan continued its seasonal cycle, while the blind box experienced a decline, potentially due to discontinuation or reduced market interest. Without knowing what the series represents, a model may struggle to capture this divergence. This illustrates a key limitation of unimodal forecasting: even with sufficient historical data, models that lack access to contextual information, such as item category, description, or external status, can misinterpret how similar-looking series actually behave. In contrast, multimodal inputs provide essential context, helping models distinguish between patterns driven by seasonality, trends, or inherent attributes.

While some recent efforts~\cite{jia2024gpt4mts,zhao2024news} have explored multimodal forecasting, existing datasets suffer from several limitations. Many are relatively small in scale~\cite{jia2024gpt4mts}, often involving fewer than 30 time series channels, which restricts their utility for training and benchmarking large models. Others, although containing time series data, are designed for domains where forecasting is not the primary task, e.g., anomaly detection, classification, clinical reasoning, or lack standardized, reproducible forecasting protocols; we discuss these in more detail in Section~\ref{sec:related}.
Additionally, some datasets focus on dynamic external modalities, such as streaming news or social media content~\cite{nyc_taxi_trip_data_icpsr,nyc_bike_sharing_zenodo2024}. 
Our work focuses on the static case, where external information like item descriptions and metadata is fixed per entity. These static additional modalities can provide additional context which can be valuable in scenarios like shown in Figure~\ref{fig:motivation}. Moreover, as shown in Section~\ref{sec:results}, these static information is essential for scenarios like cold-start forecasting where no historical time series is available. 
In this paper, we introduce a dataset suite, \textbf{Mo}dal+\textbf{Time} (\texttt{MoTime}).
\texttt{MoTime} is the largest publicly available suite of its kind, covering diverse domains and modalities. Its scale enables robust extension, training, evaluation, and generalization analysis across forecasting scenarios. Each dataset pairs time series with aligned contextual information, including but not limited to metadata, textual descriptions, categorical labels, and images. 
\texttt{MoTime} data is sourced from academically recognized sources with additional processing, including published papers and competition platforms.
We augment each time series with relevant external modalities through reframing or targeted object crawling and careful validation. 
We investigate the benefit that external modalities can bring to time series forecasting in two scenario settings. The first is \texttt{varying-history forecasting}, which analyzes when and how external modalities contribute under different lengths of historical time series in common forecasting. The second is \texttt{cold-start forecasting}, where models forecast without any prior time series history. It is an important but rarely explored setting in prior multimodal time series forecasting studies. We find that while external modalities generally enhance forecasting performance, their effectiveness varies across datasets, and the gains are notable for short series in some cases.
By framing this work as an infrastructure-level contribution, we aim to support the development of robust, context-aware forecasting models and to provide a suite for advancing multimodal research in time series modeling.

\section{Related Works}
\label{sec:related}
Recent advances in time series forecasting have spurred the release of diverse datasets and benchmarks. Compared with efforts~\cite{wang2024moirai,rasul2024lagllama,das2024decoder,chronos2024,ekambaram2024ttm,garza2025timegpt} on unimodal data~\cite{woo2024gift,nixtla2024benchmark,godahewa2021monash,dua2017uci,wang2024moirai,zhang2024moment,cohen2024toto,wikikaggle2017,zhou2024scalable,makridakis2000m3,makridakis2018m4,makridakis2022m5,zhou2021informer}, multimodal benchmarks are still in early development and are often domain-specific. We group them into general-purpose benchmarks and domain-specific datasets.

\paragraph{General-purpose multimodal benchmarks.}
Several benchmarks aim to support multimodal modeling over time series and text, though they are relatively small in scale and may not be primarily designed for forecasting.
Time-MMD~\cite{liu2024timemmd} is the first large-scale, general-purpose dataset for multimodal time series forecasting. It spans nine domains, including healthcare, finance, and agriculture, and pairs each time series with aligned dynamic textual reports. It supports multiple tasks such as forecasting, imputation, and anomaly detection, and provides an accompanying library, MM-TSFlib, for standardized evaluation. 
However, the number of channels is limited, with no domain exceeding 11 channels, see Table~\ref{tab:dataset_comparison}, which may constrain its utility for forecasting tasks.
MTBench~\cite{bai2024mtbench101} is an LLM-centric benchmark on temporal reasoning. It includes paired time series and news-style textual inputs in finance and weather domains, with tasks ranging from trend prediction and technical indicator estimation to contradiction detection and news-driven question answering. 
The benchmark emphasizes multimodal understanding over forecasting. The core evaluation targets cross-modal inference rather than standard forecasting accuracy.

These benchmarks represent important early steps toward multimodal time series understanding. However, they often focus on dynamic, event-level reasoning and offer limited support for large-scale, entity-centric forecasting with reusable protocols. Our work addresses these gaps by introducing a multimodal dataset suite with larger scales and scenario-driven forecasting tasks.

\begin{table}[htbp]
    \centering
    \caption{Comparison of multimodality time series forecasting datasets and benchmarks. TSN, TTSN, ST, and RP mean time series number, total time series number, static text, and reusable protocol.}
    \label{tab:dataset_comparison}
    \resizebox{\textwidth}{!}{
    \begin{tabular}{lcccccrcc}
    \toprule
    \textbf{Name} & \textbf{External Modality} & \textbf{Domain} & \textbf{Datasets} & \textbf{Length} & \textbf{TSN} & \textbf{TTSN} & \textbf{ST} & \textbf{RP} \\
    \midrule
    \texttt{Time-MMD} & text & climate/energy/etc & 9 & 423$\sim$11,102 & 1$\sim$11 & 36 & \xmark & \checkmark \\
    \texttt{MTBench} & text & finance/weather & 2 & 72$\sim$546~\footnotemark & 3$\sim$5 & 8 & \xmark & \checkmark \\
    \texttt{GDELT-based} & text/meta & event & 1 & $\sim$349 & 30 & 30 & \xmark & \xmark \\
    \texttt{FNF}~\cite{zhao2024news} & text & event & 5 & 1,095$\sim$70,080 & 7$\sim$862 & 913 & \xmark & \xmark \\
    \texttt{MoTime} & text/image/meta & retail/video/event & 8 & 144$\sim$10,505 & 890$\sim$86,574 & 869,466 & \checkmark & \checkmark \\
    \bottomrule
\end{tabular}
    }
\end{table}

\footnotetext{The original dataset does not specify the exact number of time steps per series, but we can approximate using typical financial settings, 7 trading days at 5-minute intervals yields roughly 546 time steps, consistent with the dataset's design.}


\paragraph{Domain-specific multimodal datasets.}
Many datasets combine time series with external modalities in specific domains. In healthcare, resources such as MIMIC-III/IV~\cite{johnson2016mimic,johnson2023mimiciv}, PTB-XL~\cite{wagner2020ptbxl}, and ICBHI~\cite{icbhi2017} integrate physiological time series with clinical reports. 
These datasets often focus on classification tasks, such as in-hospital mortality or physiological phenotyping, rather than forecasting.
In finance, datasets like FNSPID~\cite{dong2024fnspid}, GDELT-based corpora~\cite{leetaru2013gdelt,jia2024gpt4mts,zhao2024news}, and DOW30~\cite{djia2025} pair market indicators with economic narratives and event records. 
For IoT and transportation, LEMMA-RCA~\cite{zheng2024lemmarca} and NYC Taxi/Bike~\cite{nyc_taxi_trip_data_icpsr,nyc_bike_sharing_zenodo2024} combine sensor readings with spatial metadata and tags. Environmental monitoring datasets like Terra~\cite{chen2024terra} incorporate satellite-based measurements aligned with geotagged weather descriptions. 
Most use dynamic text aligned to timestamped events, which differs from static descriptions used in our setting. In addition, many lack reusable evaluation protocols and provide limited support for generalization tasks such as cold-start or entity-level forecasting. 
Most of these datasets contain a relatively small number of time series channels, typically fewer than a few dozen, as shown in Table~\ref{tab:dataset_comparison}.

\paragraph{Multimodal forecasting strategies.}
Approaches to multimodal forecasting vary in how they incorporate external modalities. Some methods transform time series into other modalities, e.g., text for language models~\cite{cao2024tempo,sun2023test,chang2023llm4ts}. However, this can disrupt the modeling of temporal dependencies. Others~\cite{jin2023time,liu2024lstprompt,wu2025dual} use shared auxiliary descriptions at the dataset or domain level, limiting granularity and flexibility in fine-grained or cold-start settings. Unlike prior work, \texttt{MoTime} is designed around static, entity-level modalities and scenario-driven tasks. It enables systematic evaluation of external modality contributions across both \texttt{short} and \texttt{long} history settings, with explicit support for generalization, cold-start, and entity-aware forecasting challenges.

\section{Data}
\label{sec:data}
We introduce \texttt{MoTime}, a suite of eight multimodal time series datasets spanning e-commerce, web traffic, media, and user behavior domains. 
\texttt{MoTime} is constructed by systematically re-purposing and transforming existing datasets, particularly from the recommender systems community, into item-centric, temporally structured forecasting tasks. 
The suite is designed to support general-purpose, multimodal forecasting and is characterized by its diversity in diverse scales, series lengths, sparsity, temporal resolution, and modality composition.
All the datasets of \texttt{MoTime} are available on \url{https://www.kaggle.com/datasets/krissssss/multimodal-time-series-forecasting/}.

\subsection{Data Overview}

\texttt{MoTime} consists of eight datasets, organized into two main categories based on their origin: (1) recommender system datasets re-purposed for forecasting, and (2) web and media popularity datasets. We briefly describe each dataset below; further details, including sources, are provided in Appendix~\ref{sec:dataset_details}.

\paragraph{Recommender datasets transformed into time series forecasting.}
These datasets are originally designed for personalized ranking or click-through prediction. We convert user-item interactions into item-oriented time series that reflect popularity dynamics over time.

\texttt{PixelRec}~\cite{cheng2023image} captures short video behavior in lifestyle and entertainment domains. We aggregate user interactions into daily view series. Each item includes a thumbnail and title metadata. The series is long and sparse.
\texttt{TaobaoFashion}\footnote{\url{https://tianchi.aliyun.com/competition/entrance/231506/information}} provides outfit-level purchase logs. We construct daily purchase series per item, each paired with an image, making it well-suited for short-horizon, image-conditioned forecasting.
\texttt{AmazonReview} consists of item-level review logs across 29 categories. We derive daily review count series per item and align them with item metadata such as title, description, category, and price. The data supports semantic-aware forecasting.
\texttt{Tianchi}\footnote{\url{https://tianchi.aliyun.com/dataset/43}} offers large-scale purchase behavior from an e-commerce platform. We extract item-level purchase series and align them with both text fields and images, enabling trend-aware, multimodal forecasting.
\texttt{MovieLens} is a classical benchmark in recommendation. We convert rating logs into daily interaction series per movie and enrich them with externally extracted metadata, e.g., overview, genres, and tags. It serves as a benchmark for sparse, text-enhanced forecasting.

\paragraph{Media and web traffic datasets.}
These datasets naturally contain time series aligned with content metadata and reflect the dynamics of social or online attention.
\texttt{News}~\cite{moniz2018news} captures early popularity of news articles over 48 intervals at 20-minute resolution. Each article includes headline text, topic, and sentiment scores, making it ideal for fine-grained, multimodal attention forecasting.
\texttt{WikiPeople}~\cite{yang2017web} includes multichannel daily view counts, e.g., desktop, mobile, for person-related articles. Text summaries are aligned with article IDs, supporting cross-device and text-conditioned forecasting.

\paragraph{Additional dataset.}
\texttt{VISUELLE}~\cite{Skenderi_2022_CVPR} provides visual and temporal engagement on Instagram posts. It includes image posts along with associated metadata such as tags, captions, and engagement statistics over time. The time series is item sales. While we do not process or experiment on VISUELLE in this work, we include it in the \texttt{MoTime} collection for future benchmarking.

\begin{table}[ht]
    \centering
    \caption{Statistics of the eight multimodal time series datasets in \texttt{MoTime}.}
    \label{tab:dataset_stats}
    \resizebox{\textwidth}{!}{%
    \begin{tabular}{lcrcccl}
    \toprule
    \textbf{Dataset} & \textbf{TS Shape} & \textbf{Density(\%)} & \textbf{Text} & \textbf{Image} & \textbf{Metadata} & \textbf{Notes} \\
    \midrule
    PixelRec & 4,865 $\times$ 43,082 & 4.41 & \checkmark & \checkmark & \checkmark & Long sparse multivariate TS \\
    TaobaoFashion & 365 $\times$ 890 & 68.01 & -- & \checkmark & -- & One image per item \\
    MovieLens & 10,505 $\times$ 84,518 & 1.66 & \checkmark & -- & \checkmark & Text scraped externally \\
    AmazonReview & 3,934 $\times$ 668,756 & 6.18 & \checkmark & -- & \checkmark & 29 categories, sparse TS \\
    Tianchi & 365 $\times$ 36,397 & 53.15 & \checkmark & \checkmark & -- & E-commerce purchase logs \\
    News & 144 $\times$ 26,612 & 17.61 & \checkmark & -- & \checkmark & 20-min interval resolution \\
    WikiPeople & 550 $\times$ 3,856 & 99.96 & \checkmark & -- & \checkmark & Multichannel access modes \\
     VISUELLE &  11 $\times$ 5,355  & 62.48 & \checkmark & \checkmark & \checkmark & Irregular time series\\
    \bottomrule
    \end{tabular}}
\end{table}

\subsection{Data Construction and Processing}
In this section, we summarize the key steps in processing the data. More details are provided in Appendix~\ref{sec:dataset_details}.
To construct \texttt{MoTime}, we transform source datasets into time-indexed series aligned with external modalities. This involves three key steps: (1) generating time series from raw interaction or popularity data, (2) extracting and aligning modality information, and (3) filtering and cleaning to ensure consistency and usability.
\textbf{Time series construction.}
For datasets originating from recommender systems, \texttt{PixelRec}, \texttt{TaobaoFashion}, \texttt{Tianchi}, and \texttt{AmazonReview}, we convert raw user-item interactions into item-centric time series. Specifically, we aggregate user behaviors into daily-level popularity signals for each item. This reframing enables classical recommendation data to be used in forecasting settings, where the task is to predict how item popularity evolves over time. In \texttt{MovieLens}, we similarly aggregate rating histories into daily interaction series for each movie. For \texttt{News}, \texttt{WikiPeople}, the time series are directly available as view or engagement counts at regular time intervals.
\textbf{Modality extraction and alignment.}
External modalities, including textual descriptions, item images, and structured metadata, are either extracted from the original datasets or retrieved externally. For \texttt{PixelRec}, \texttt{TaobaoFashion}, \texttt{Tianchi}, and \texttt{AmazonReview}, image and/or text features are already provided. We align these with time series using unified item identifiers defined within each dataset. In contrast, for \texttt{MovieLens} and \texttt{WikiPeople}, we obtain external text by crawling movie metadata or Wikipedia summaries, and link them to the corresponding series using consistent IDs.
\textbf{Filtering and cleaning.}
During preprocessing, we remove corrupted or incomplete entries from each modality. For textual data, we discard samples with missing or invalid fields. For image data, we retain only samples that can be reliably linked to time series objects. For \texttt{MovieLens}, we filter out series that are too short or have too many sparse observations. All retained samples have fully aligned time series and modality information, ensuring that multimodal learning can be conducted in a consistent and reproducible way.
\textbf{Cold-start support.}
Three datasets, \texttt{AmazonReview}, \texttt{MovieLens}, and \texttt{News}, include explicit release or publication timestamps for each item. This allows us to identify time steps prior to an item’s availability and annotate them accordingly. These pre-release segments enable the design of cold-start forecasting tasks, where models must make predictions based solely on external modality signals.

\subsection{Statistical Summary}
Table~\ref{tab:dataset_stats} presents a unified summary of eight datasets in \texttt{MoTime}, detailing their time series scale including obervations per time series and the number of series per dataset, data density, and modality composition. Density is reported as a percentage, offering insight into sparsity levels and potential cold-start scenarios. Additional statistics of time series, including per-series mean, median, and value range, are included in Table~\ref{tab:ts_stats}, and statistics of text, refer to~\ref{tab:ts_stats} in Appendix.

The datasets exhibit substantial diversity in scale and channels. 
For instance, \texttt{PixelRec} contains 43,082 long, sparse series, while \texttt{TaobaoFashion} includes 890 short, dense series aligned with item images. 
\texttt{WikiPeople} offers 3,856 dense (99.96\%), multichannel time series, making it well-suited for standard forecasting setups. In contrast, \texttt{AmazonReview} consists of 29 category-specific sub-datasets, each with its own semantic domain and sparsity level. This hierarchical structure allows for the evaluation of domain transfer, few-shot generalization, and model robustness under distribution shift.
\texttt{News} captures the short-term popularity of news articles driven by real-world events. It is the only dataset in \texttt{MoTime} where series are aligned by relative time, with each series starting from the moment of publication and covering a high temporal resolution. Making the dataset particularly useful for evaluating models on rapid trend emergence, early-stage signal detection, and time-lagged multimodal influence.

Modality configurations are also diverse. \texttt{MovieLens}, \texttt{AmazonReview}, and \texttt{WikiPeople} are text-only; \texttt{TaobaoFashion} is image-only; while \texttt{Tianchi}, \texttt{PixelRec},  are fully multimodal. All modalities are aligned with time series via consistent sample IDs.
Taken together, MoTime supports a broad spectrum of forecasting tasks and scenarios, from fine-grained modeling to cold-start forecasting to semantic long-range forecasting. Its diversity in sparsity, series length, and modality alignment enables robust benchmarking for both unimodal and multimodal forecasting models.

\section{Multimodal Utility under Different Forecasting Scenarios}
To comprehensively explore when and how external modalities benefit forecasting performance, we design two representative scenarios: varying-history and cold-start forecasting.
These scenarios reflect practical challenges in real-world forecasting applications and are rarely addressed systematically in prior benchmarks.

\subsection{Modality Utility under Scenario 1: Varying-history Forecasting}
\label{sec:scenario1}
The motivation for this scenario stems from a hypothesis: additional information is most beneficial when the time series itself provides limited signals, and becomes less critical when the temporal signal is already strong. In particular, we hypothesize that \texttt{short} history series are more reliant on external modalities, while \texttt{long} history series may already contain sufficient temporal patterns for accurate forecasting. To test this hypothesis across diverse domains, we construct a training setup that explicitly contrasts both \texttt{short} history and \texttt{long} history inputs, enabling us to evaluate the marginal contribution of modalities under varying temporal availability under common forecasting.

We randomly split the training set into two subsets: one retaining the full historical windows (\texttt{long}), and one using only a few of each series (\texttt{short}). The model is trained jointly on both subsets, while the validation and test sets remain unchanged to ensure comparability. This setup allows for consistent evaluation of modality effectiveness under both sufficient history and limited-history conditions. 

As our proposed baseline method for forecasting in this scenario, we adopt a dual-tower architecture inspired by TextFusionHTS~\cite{xin2024textfusionhts}, comprising a time series encoder and a frozen LLM encoder. Image inputs are first converted into captions via a vision-language model and processed alongside text using the same encoder. The modality-specific representations are concatenated and passed through a lightweight MLP for final forecasting.
The embedding-based architecture of \texttt{PatchTST}~\cite{nie2023time} aligns well with multimodal fusion and serves as a candidate for extension in multimodalities. Thus, we adapt it to integrate the multimodal information and propose \texttt{MultiPatchTST}.
While recent state-of-the-art \texttt{WPMixer}~\cite{murad2024wpmixer} originally unimodal, we adapt it to \texttt{MultiWPMixer} by incorporating contextual embeddings to decomposed components.

\subsection{Modality Utility under Scenario 2: Cold-start Forecasting} 
The forecast of cold-start forecasting must rely entirely on the entity's time-irrelevant external modalities, where no historical time series is available. It remains underexplored due to data and evaluation limitations.
One key aim of \texttt{MoTime} is to enable systematic investigation of this challenging setting, which allows us to assess the utility of external modalities and highlight the potential of multimodal information under data sparsity.

As the proposed baseline method, we adopt a retrieval-augmented generation pipeline inspired by a recent work on cold-start web traffic forecasting~\cite{zhou2024ccwtf}. Specifically, we first construct a retrieval base composed of existing time series instances and their associated metadata, textual descriptions, or image captions. All external modalities are embedded with an LLM, and the resulting vectors are cached for fast retrieval. This retrieval base serves as a semantic index to support cold-start inference.
At inference time, the input for a new entity is its textual description only. We encode this text into an embedding using the same frozen LLM model. Cosine similarity is then computed between the input embedding and all stored embeddings in the retrieval base to obtain a relevance score for each candidate. We retain the top-$k$ most similar entities and retrieve both their text and corresponding time series data.
The selected top-$k$ time series are aligned with their corresponding text descriptions to form a structured input prompt. This prompt is then fed into a generation model to produce the forecast for the cold-start entity. The model is conditioned on both retrieved trajectories and metadata, enabling cold-startforecasting purely based on semantic analogy.
We then construct a structured prompt containing: (1) the textual description of the target entity, (2) textual descriptions of the $k$ most similar entities (including converted image descriptions), (3) the historical time series of these $k$ relevant entities, and (4) their corresponding entity IDs and timestamp information. This prompt is passed to an LLM, which is tasked with generating a forecast for the target entity. To ensure structured output and improve reasoning consistency, we constrain the output format and instruct the model to provide reasoning behind its prediction. For more details about the prompt, please refer to Table~\ref{tab:coldstart_prompt} in Appendix.

\section{Experiment}
We evaluate forecasting performance across two scenarios enabled by \texttt{MoTime}.
\subsection{Evaluation Setup}\label{sec:evaluation}
\paragraph{Baselines.}
In varying-history forecasting, we consider several representative baselines to cover different modeling paradigms. 
\texttt{DLinear}~\cite{zeng2022are}: A linear decomposition-based model that separately maps components to forecast values. As it avoids latent embeddings, DLinear does not naturally extend to multimodal variants and is included in its original form.
\texttt{PatchTST}~\cite{nie2023time}: A Transformer-based model that operates on patch embeddings of time series inputs. 
\texttt{WPMixer}~\cite{murad2024wpmixer}: A recent state-of-the-art model that applies learned filters to decomposed components, followed by lightweight feed-forward layers. In this varying-history forecasting, we compare these baselines with our proposed \texttt{MultiPatchTST} and \texttt{MultiWPMixer} as introduced in Section \ref{sec:scenario1}.
Given the limited attention to cold-start forecasting in existing literature and fewer multimodal resources, we use a simple baseline: the \texttt{average} of retrieved relevant series, computed independently for each forecast horizon. This baseline provides a robust point of comparison for evaluating the benefit of textual retrieval and generation-based modeling.
As a baseline, we independently compute the retrieved series' element-wise average for each forecast horizon. This non-parametric baseline provides a strong reference point for assessing the added value of generation-based forecasting conditioned on semantic context.

\paragraph{Metrics.} 
We report two widely used metrics~\cite{godahewa2021monash},
\begin{equation*}
    \text{RMSE} = \sqrt{\frac{1}{T} \sum_{t=1}^{T} (y_t - \hat{y}_t)^2}, \quad
    \text{WRMSPE} = \frac{\sqrt{\frac{1}{T} \sum{t=1}^{T} (y_t - \hat{y}_t)^2}}{\frac{1}{T} \sum{t=1}^{T} |y_t|}
\end{equation*}
RMSE captures error in the original scale and is particularly sensitive to large deviations.
WRMSPE normalizes RMSE by the mean absolute value of the ground truth, offering a scale-invariant view of forecasting quality.
We focus on squared-error metrics as they are sensitive to large deviations, which is important in sparse/intermittent or spiky series, a property that many series in our datasets have. We note that we do not use scaled measures that are popular in forecasting contexts, such as the RMSSE, because our test spans are long enough to compute meaningful absolute-scale metrics over the test sets. Moreover, RMSSE becomes difficult to interpret when forecast horizons vary across test cases, as in our setting.

\paragraph{Protocols.} 
In varying-history forecasting, we split the train, validate, and test set as the ratio of 7:1:2 of the longest series. To mimic varying lengths, we split the series into \texttt{long} and \texttt{short} groups at a 1:1 ratio. The threshold for defining \texttt{short} history series is determined based on the dataset’s temporal resolution and the typical length of its series: 100 steps for \texttt{PixelRec}, \texttt{AmazonReview}, and \texttt{WikiPeople}, 50 for \texttt{Movielens}, 20 for \texttt{TaobaoFashion} and \texttt{Tianchi}, and 18 for \texttt{News}. These values reflect meaningful cutoffs for varying-history forecasting in each domain.
Models are trained with both groups, enabling them to generalize across heterogeneous history lengths. We evaluate forecasting performance on the \texttt{short}, \texttt{long} series, and \texttt{mixture} of both separately. Input and output lengths are adapted to dataset frequency: daily datasets use 7-day inputs and forecast 7 to 28 days ahead; the high-frequency dataset, \texttt{News}, uses 6-step inputs and forecasts up to 12 steps ahead.
All reported scores are computed on the original scale of the data without normalization or scaling being applied before evaluation. This choice preserves the meaningfulness of errors. Consequently, datasets with inherently large values, e.g., \texttt{WikiPeople}, \texttt{News}, and \texttt{Tianchi}, exhibit correspondingly large RMSE values.

In cold-start setting, we randomly sample 30 series as cold-starters. Forecasting starts from the first valid time step, i.e., non-zero and non-placeholder value, using only 7 previous daily steps or 6 previous 20-minute steps from the relevant, non-target entity, depending on the dataset frequency.

\paragraph{Implementation details.} \label{sec:implement}
In varying-history forecasting, we apply reindexing~\cite{zhou2024scalable} to ensure mixed batch samples of \texttt{short} and \texttt{long}.
To achieve a diverse experiment, we filter out the series whose density is less than 0.4 in \texttt{PixelRec}, and randomly sample 1,000 series from \texttt{MovieLens}.
In cold-start forecasting, we simulate cold-start conditions by randomly selecting 30 channels from each dataset and removing all but the first valid observation, excluding 0s and -1s. For each target, we retrieve the top 4 relevant series based on textual similarity, using GPT-4o-mini embeddings as retrievers. Forecasting is then performed using a GPT-4o-mini model conditioned on the relevant series and metadata.
Models are trained with MSE loss using the Adam optimizer. Time series encoders and modality fusion layers are updated; contextual encoders remain fixed. Early stopping is based on validation loss.
All data processing and experiments are conducted on a single GPU (NVIDIA A100, A40, or RTX 3090), selected based on availability. The overall setup is designed to be reproducible and computationally feasible on commonly available hardware.
For more computation information, e.g., running time, please refer to Appendix~\ref{appendix:implement}.

\begin{table}[htbp]
    \centering
    \caption{Evaluation on varying-training length forecasting (\texttt{long} history series). Due to space limitations, evaluation scores are rounded to three decimal places. In cases where multiple models appear to have identical scores under this rounding, we still mark the best and second-best results in bold and with an underline based on the full-precision metrics.}
    \label{tab:long_results_2}
    \resizebox{\textwidth}{!}{
    \begin{tabular}{l|r|rr|rr|rr|rr|rr}
        \hline
         \multicolumn{2}{c|}{Models} & \multicolumn{2}{c|}{DLinear} & \multicolumn{2}{c|}{PatchTST} & \multicolumn{2}{c|}{WPMixer} & \multicolumn{2}{c|}{MultiPatchTST} & \multicolumn{2}{c}{MultiWPMixer} \\
         \hline
         \multicolumn{2}{c|}{Metric} & rmse & wrmspe & rmse & wrmspe & rmse & wrmspe & rmse & wrmspe & rmse & wrmspe \\
         \toprule
         \multirow{5}{*}{\rotatebox{90}{PixelRec}} & 1 & 1.379 & 1.606 & \underline{1.322} & \underline{1.541} & 1.356 & 1.580 & 1.328 & 1.548 & \textbf{1.315} & \textbf{1.532} \\
        & 7 & 1.547 & 1.798 & 1.444 & 1.678 & 1.456 & 1.692 & \textbf{1.427} & \textbf{1.658} & \underline{1.439} & \underline{1.672} \\
        & 14 & 1.577 & 1.824 & 1.501 & 1.736 & 1.509 & 1.746 & \textbf{1.484} & \textbf{1.716} & \underline{1.496} & \underline{1.730} \\
        & 21 & 1.647 & 1.897 & 1.546 & 1.781 & 1.551 & 1.787 & \textbf{1.529} & \textbf{1.761} & \underline{1.541} & \underline{1.775} \\
        & 28 & 1.648 & 1.891 & 1.584 & 1.817 & 1.593 & 1.828 & \textbf{1.570} & \textbf{1.801} & \underline{1.577} & \underline{1.809} \\
         \hline
         \multirow{5}{*}{\rotatebox{90}{Amazon}} & 1 & \textbf{0.373} & \textbf{3.604} & 0.374 & 3.620 & 0.377 & 3.649 & \underline{0.373} & \underline{3.612} & 0.376 & 3.642 \\
        & 7 & 0.394 & 3.797 & 0.390 & 3.763 & 0.390 & 3.765 & \textbf{0.389} & \textbf{3.750} & \underline{0.390} & \underline{3.763} \\
        & 14 & 0.403 & 3.873 & 0.401 & 3.851 & \underline{0.400} & \underline{3.844} & \textbf{0.400} & \textbf{3.838} & 0.401 & 3.848 \\
        & 21 & 0.411 & 3.937 & 0.409 & 3.909 & \underline{0.408} & \underline{3.906} & \textbf{0.407} & \textbf{3.898} & 0.409 & 3.911 \\
        & 28 & 0.417 & 3.974 & 0.415 & 3.955 & \underline{0.414} & \underline{3.949} & \textbf{0.414} & \textbf{3.945} & 0.415 & 3.956 \\
         \hline
         \multirow{5}{*}{\rotatebox{90}{Taobao}} & 1 & 6.203 & 1.808 & \underline{5.864} & \underline{1.709} & 6.414 & 1.869 & 6.177 & 1.800 & \textbf{5.632} & \textbf{1.641} \\
        & 7 & 6.228 & 1.814 & \underline{6.073} & \underline{1.768} & \textbf{6.066} & \textbf{1.767} & 6.313 & 1.838 & 6.155 & 1.793 \\
        & 14 & 7.066 & 2.047 & \textbf{6.302} & \textbf{1.826} & 6.496 & 1.882 & 6.405 & 1.856 & \underline{6.375} & \underline{1.847} \\
        & 21 & 7.107 & 2.058 & 6.618 & 1.916 & 6.749 & 1.954 & \textbf{6.613} & \textbf{1.915} & \underline{6.592} & \underline{1.909} \\
        & 28 & 7.289 & 2.113 & 6.879 & 1.994 & 6.974 & 2.021 & \textbf{6.839} & \textbf{1.982} & \underline{6.853} & \underline{1.986} \\
         \hline
         \multirow{3}{*}{\rotatebox{90}{Tianchi}} & 1 & 186.20 & 11.42 & 184.10 & 11.29 & 185.50 & 11.37 & \underline{180.50} & \underline{11.07} & \textbf{179.98} & \textbf{11.03} \\
        & 7 & 181.70 & 11.43 & 175.40 & 11.03 & 176.40 & 11.09 & \underline{174.60} & \underline{10.98} & \textbf{174.28} & \textbf{10.96} \\
        & 14 & 189.00 & 12.17 & \textbf{175.80} & \textbf{11.32} & 176.00 & 11.33 & \underline{175.90} & \underline{11.33} & 176.02 & 11.34 \\
         \hline
         \multirow{5}{*}{\rotatebox{90}{Movielens}} & 1 & \textbf{0.260} & \textbf{6.030} & 0.262 & 6.073 & \underline{0.260} & \underline{6.032} & 0.262 & 6.079 & 0.261 & 6.043 \\
         & 7 & \textbf{0.265} & \textbf{6.141} & 0.268 & 6.213 & 0.267 & 6.172 & 0.274 & 6.338 & \underline{0.266} & \underline{6.164} \\
         & 14 & \textbf{0.267} & \textbf{6.189} & 0.274 & 6.332 & \underline{0.269} & \underline{6.233} & 0.279 & 6.459 & 0.271 & 6.282 \\
         & 21 & \textbf{0.269} & \textbf{6.229} & 0.279 & 6.466 & \underline{0.272} & \underline{6.303} & 0.286 & 6.624 & 0.275 & 6.360 \\
         & 28 & \textbf{0.271} & \textbf{6.274} & 0.283 & 6.542 & \underline{0.273} & \underline{6.307} & 0.290 & 6.717 & 0.277 & 6.406 \\
         \hline
         \multirow{5}{*}{\rotatebox{90}{News}} & 1 & 57.43 & 47.82 & \textbf{61.33} & \textbf{51.07} & 58.39 & 48.62 & 58.89 & 49.05 & \underline{57.89} & \underline{48.21} \\
        & 3 & 60.02 & 47.42 & 60.30 & 47.63 & \textbf{59.37} & \textbf{46.90} & 59.87 & 47.29 & \underline{59.48} & \underline{46.99} \\
        & 6 & 51.41 & 40.83 & 51.52 & 40.91 & \underline{51.06} & \underline{40.55} & 51.20 & 40.66 & \textbf{51.06} & \textbf{40.55} \\
        & 9 & 46.56 & 36.41 & 46.54 & 36.39 & \underline{46.18} & \underline{36.10} & 46.42 & 36.29 & \textbf{46.15} & \textbf{36.08} \\
        & 12 & 44.86 & 33.46 & 44.64 & 33.29 & \underline{44.35} & \underline{33.08} & 44.48 & 33.18 & \textbf{44.31} & \textbf{33.05} \\
         \hline
         \multirow{5}{*}{\rotatebox{90}{WikiPeople}} & 1 & 21977 & 5.105 & \underline{20767} & \underline{4.824} & 20797 & 4.831 & 20844 & 4.841 & \textbf{20651} & \textbf{4.796} \\
         & 7 & 22174 & 5.235 & \textbf{21571} & \textbf{5.093} & 21635 & 5.108 & \underline{21583} & \underline{5.096} & 21596 & 5.099 \\
         & 14 & 19860 & 4.741 & \underline{18527} & \underline{4.423} & 18563 & 4.432 & \textbf{18519} & \textbf{4.421} & 18568 & 4.433 \\
         & 21 & 18684 & 4.441 & \underline{17666} & \underline{4.199} & 17754 & 4.220 & \textbf{17646} & \textbf{4.194} & 17699 & 4.206 \\
         & 28 & 19063 & 4.488 & \underline{17493} & \underline{4.119} & 17588 & 4.141 & \textbf{17461} & \textbf{4.111} & 17562 & 4.135 \\
         \bottomrule
    \end{tabular}
    }
\end{table}

\subsection{Results and Analysis}
\label{sec:results}
This section presents key observations under the two scenarios. Rather than emphasizing specific model performance, we focus on how different datasets and scenarios enable systematic evaluation of multimodal forecasting.

\subsubsection{Varying-history Forecasting} 
While we initially hypothesized that external modalities would be especially beneficial for short series~\ref{sec:scenario1}, the empirical results show that this hypothesis holds only in specific datasets, rather than universally. Across most datasets, we find that the performance trends, in terms of model rankings, multimodal benefits, and horizon-specific performance, across \texttt{short} and \texttt{long} subsets, are nearly consistent. Thus, we present the main results for \texttt{long} series in Table~\ref{tab:long_results_2} and \texttt{short} series and \texttt{mixture} results in the Appendix~\ref{sec:appresults}.
However, one notable exception: on \texttt{Movielens}, the short subset sees a better improvement from multimodal input with \texttt{MultiWPMixer}. This may be attributed to the sparsity of interaction data in short series, where textual context compensates for the lack of temporal signal. The gain may also stem from \texttt{WPMixer}’s ability to capture local decomposed structure, which aligns well with the lightweight signals in sparse data.

There are some interesting observations based on Table~\ref{tab:long_results_2}. \texttt{WikiPeople}, \texttt{AmazonReview}, and \texttt{News} all exhibit stronger multimodal gains at longer horizons, suggesting that external information becomes increasingly valuable as the temporal signal fades or becomes more uncertain. \texttt{PixelRec} shows clear benefits from multimodal input, indicating that item-level static text is important role when series are relatively long.
\texttt{TaobaoFashion} displays a horizon-dependent pattern: for short horizons, the best performance comes from the \texttt{WPMixer}-based multimodal model, \texttt{MultiWPMixer}; for longer horizons, the \texttt{PatchTST}-based multimodal model, \texttt{MultiPatchTST}, takes the lead. This may reflect different temporal features captured by linear and transformer.
\texttt{Tianchi} shows stronger multimodal gains at shorter horizons, possibly because short-range dynamics are more sensitive to context across items, while longer-range trends are more stable and thus less reliant on external context.

Overall, these results suggest that the utility of external modalities depends not only on series length but also on data sparsity, forecast horizon, and modality alignment. \texttt{MoTime} enables systematic analysis of these interactions, rather than assuming uniform modality contributions across tasks.

\subsubsection{Cold-start Forecasting}
\begin{figure}[htbp]
    \centering
    \includegraphics[width=0.14\textwidth]{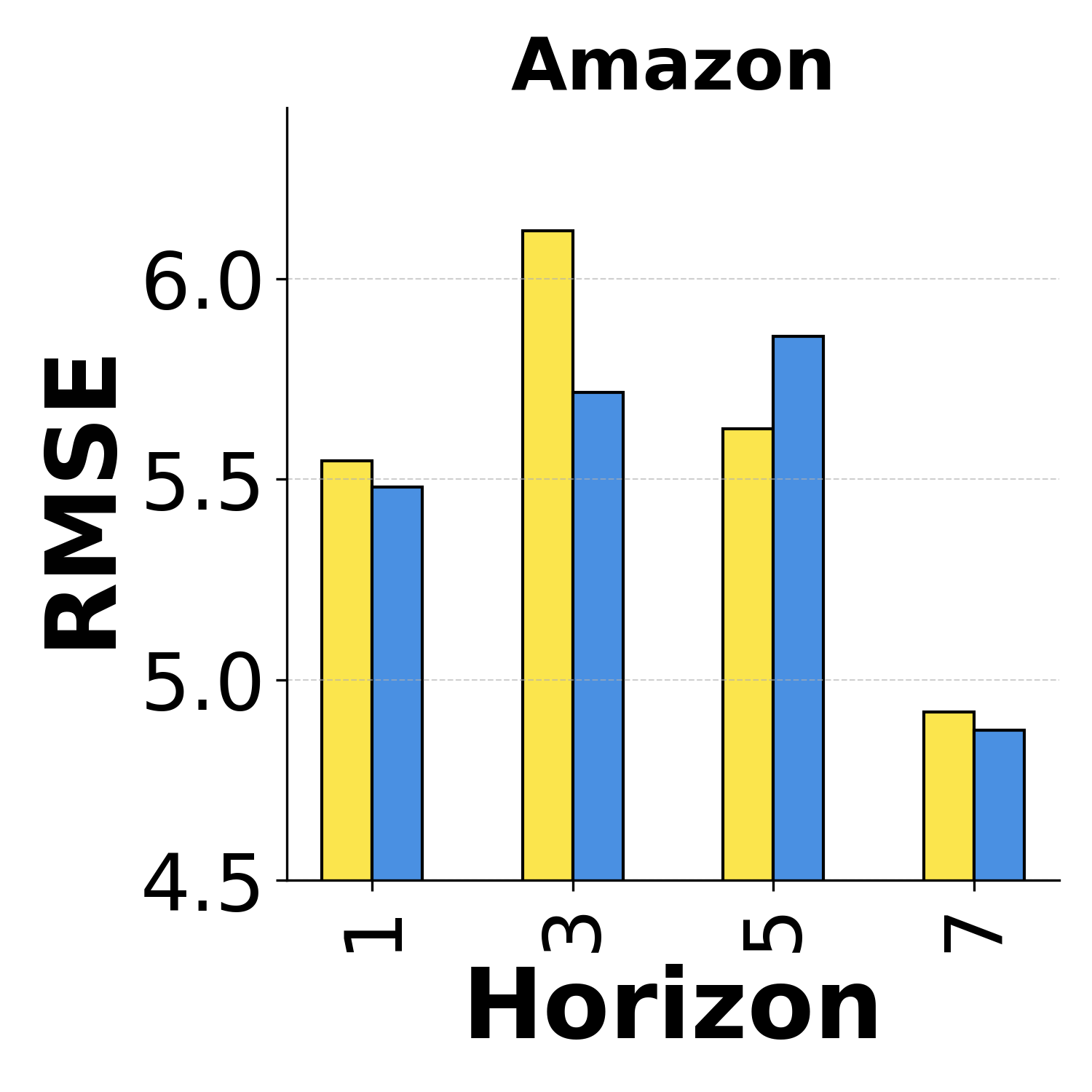}\hspace{2mm}
    \includegraphics[width=0.14\textwidth]{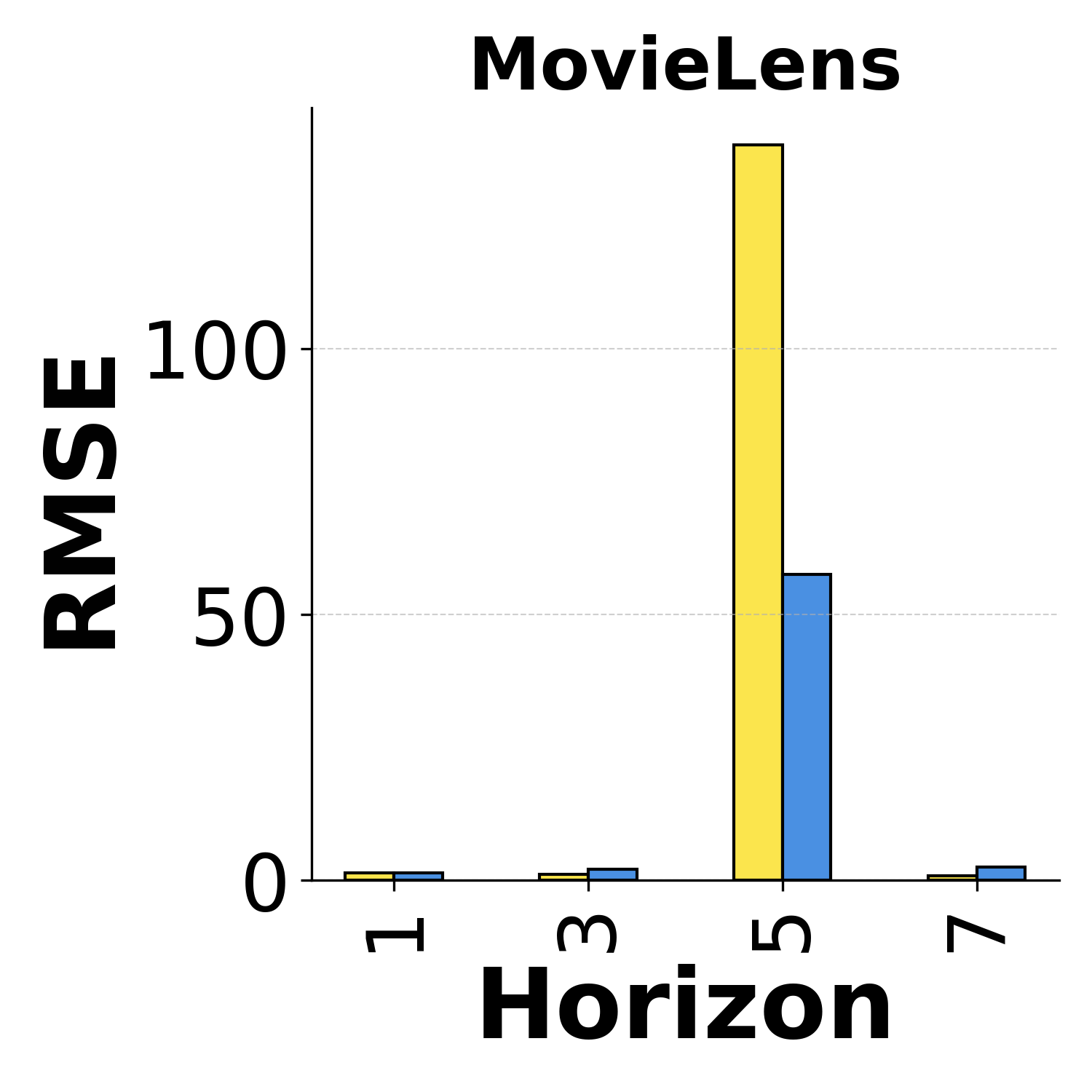}\hspace{2mm}
    \includegraphics[width=0.14\textwidth]{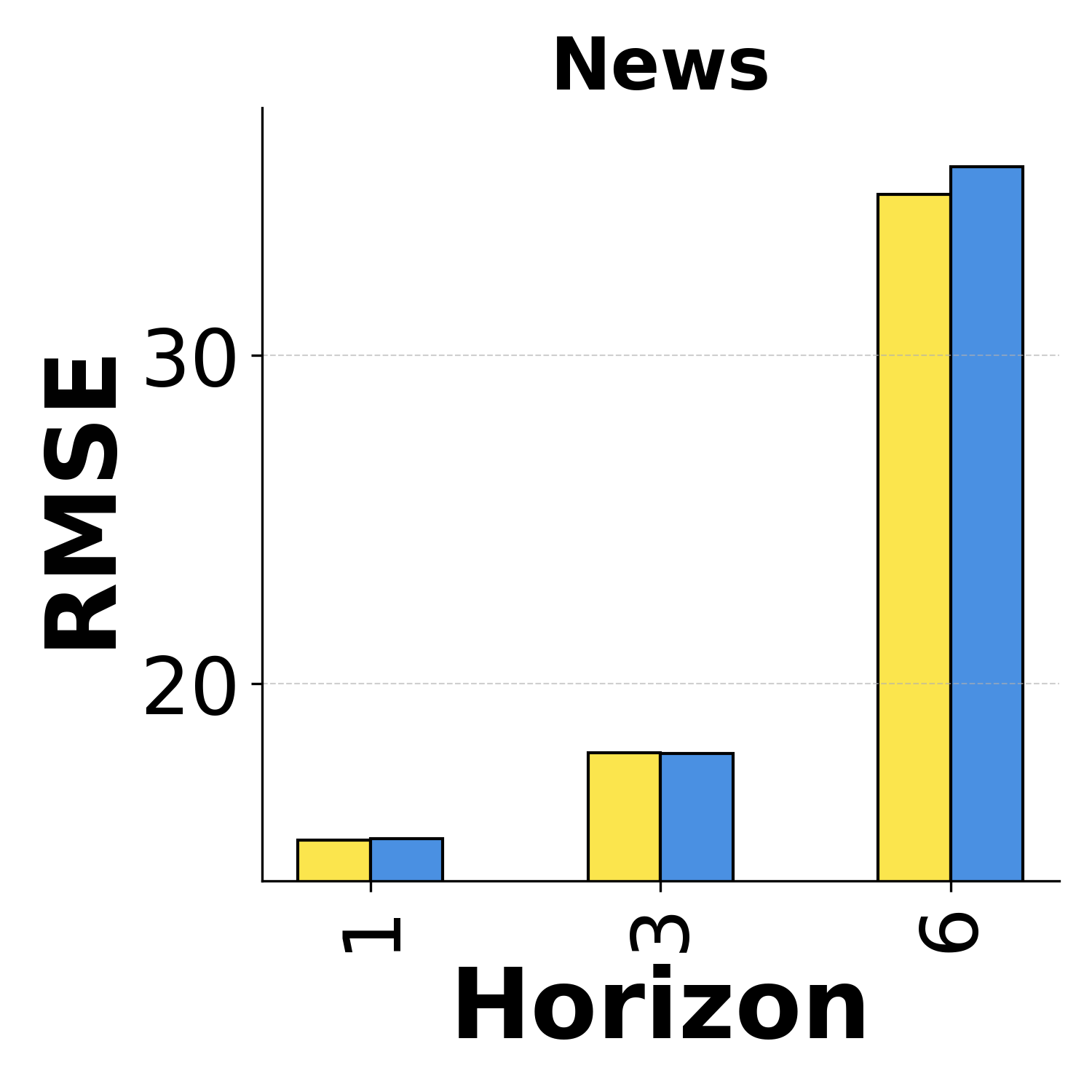}\hspace{2mm}
    \includegraphics[width=0.14\textwidth]{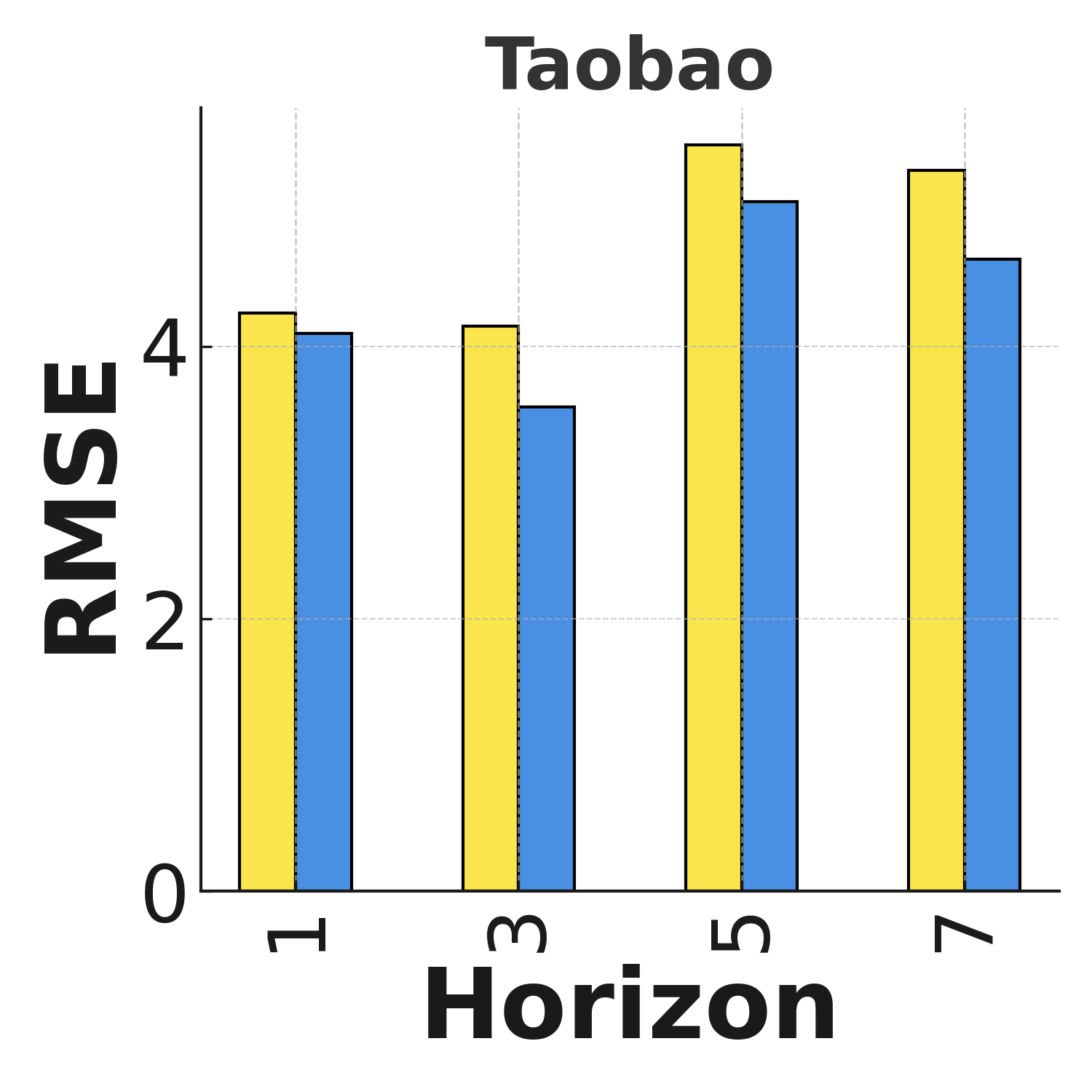}\hspace{2mm}
    \includegraphics[width=0.14\textwidth]{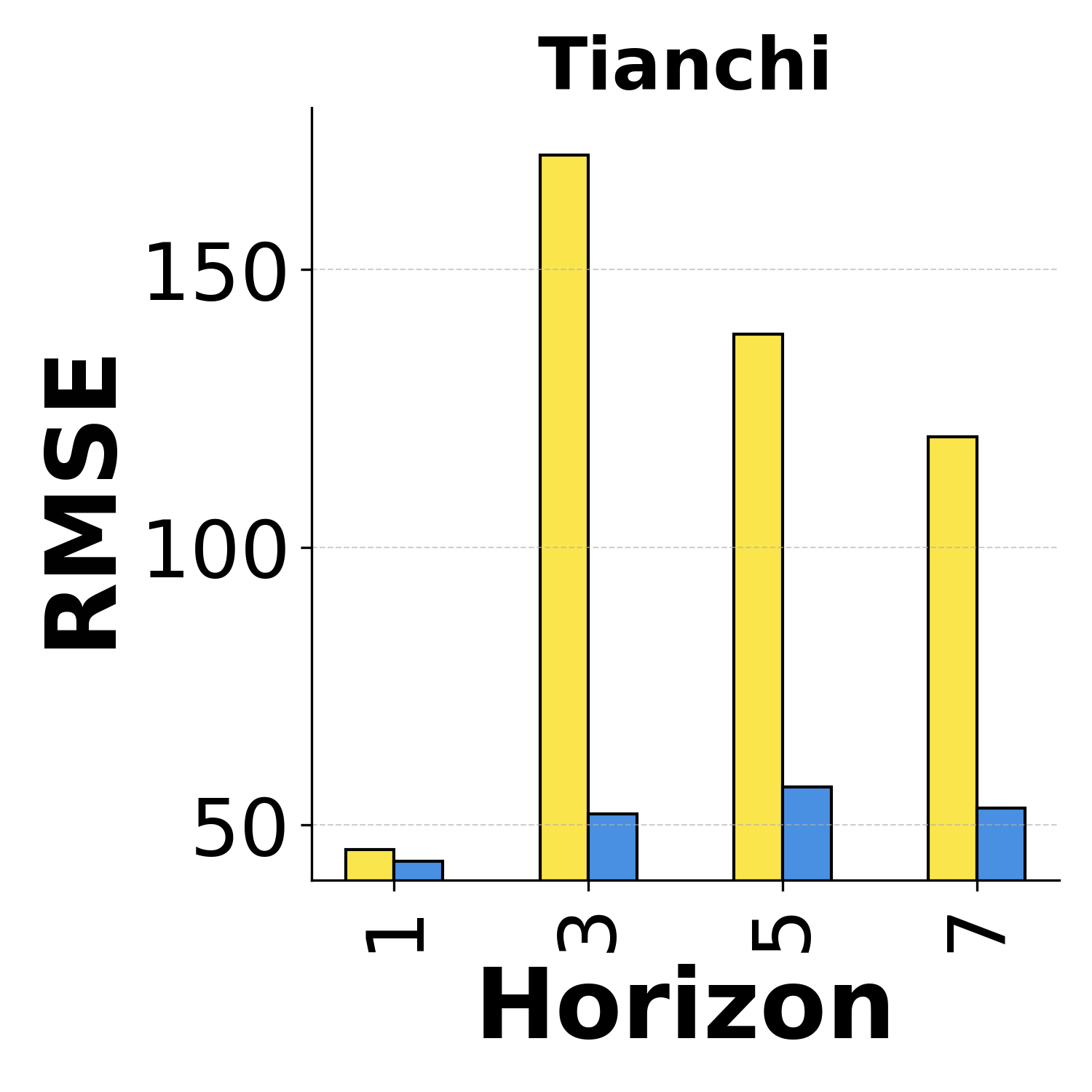}\hspace{2mm}
    \includegraphics[width=0.14\textwidth]{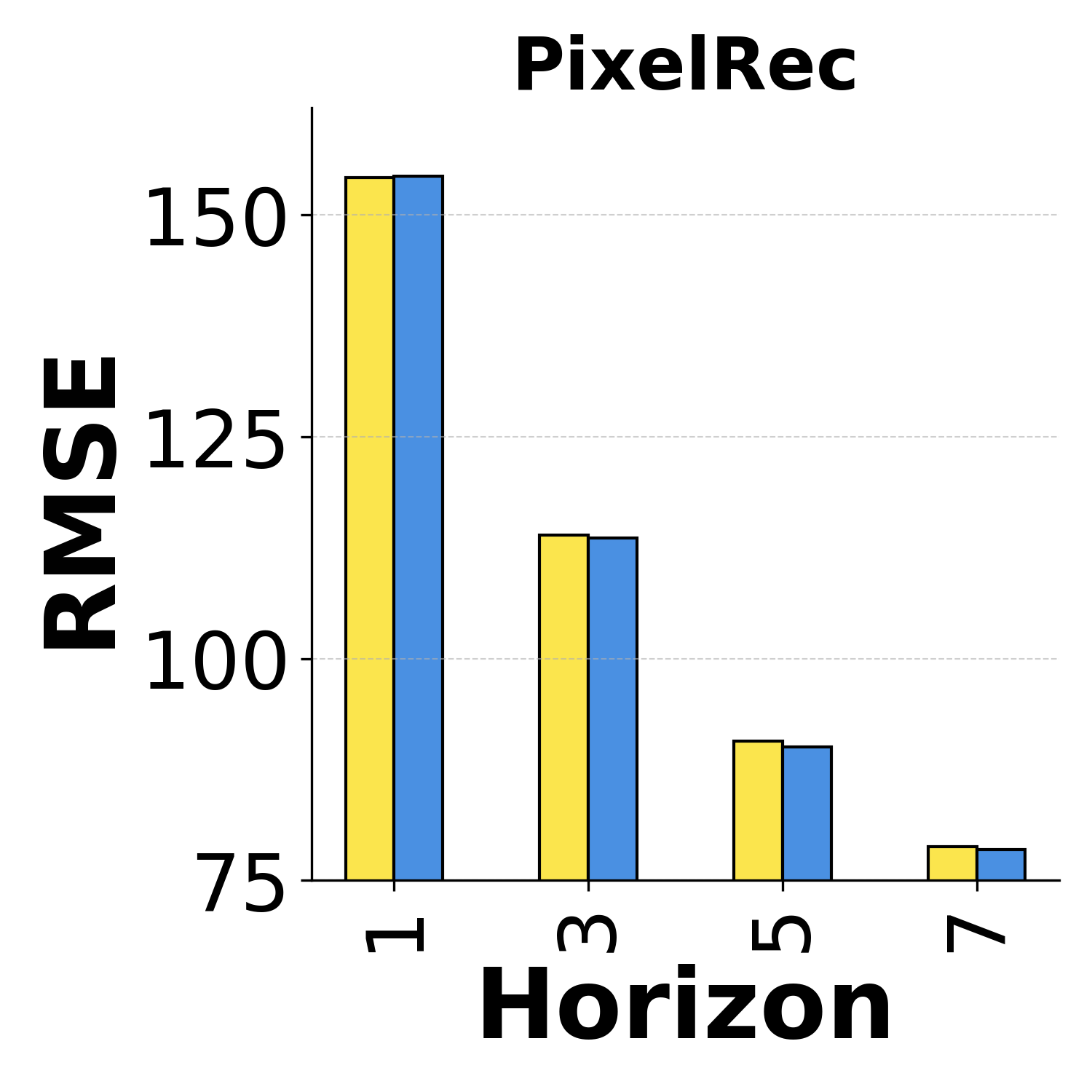}
    
    \vspace{0mm} %
    
    \includegraphics[width=0.14\textwidth]{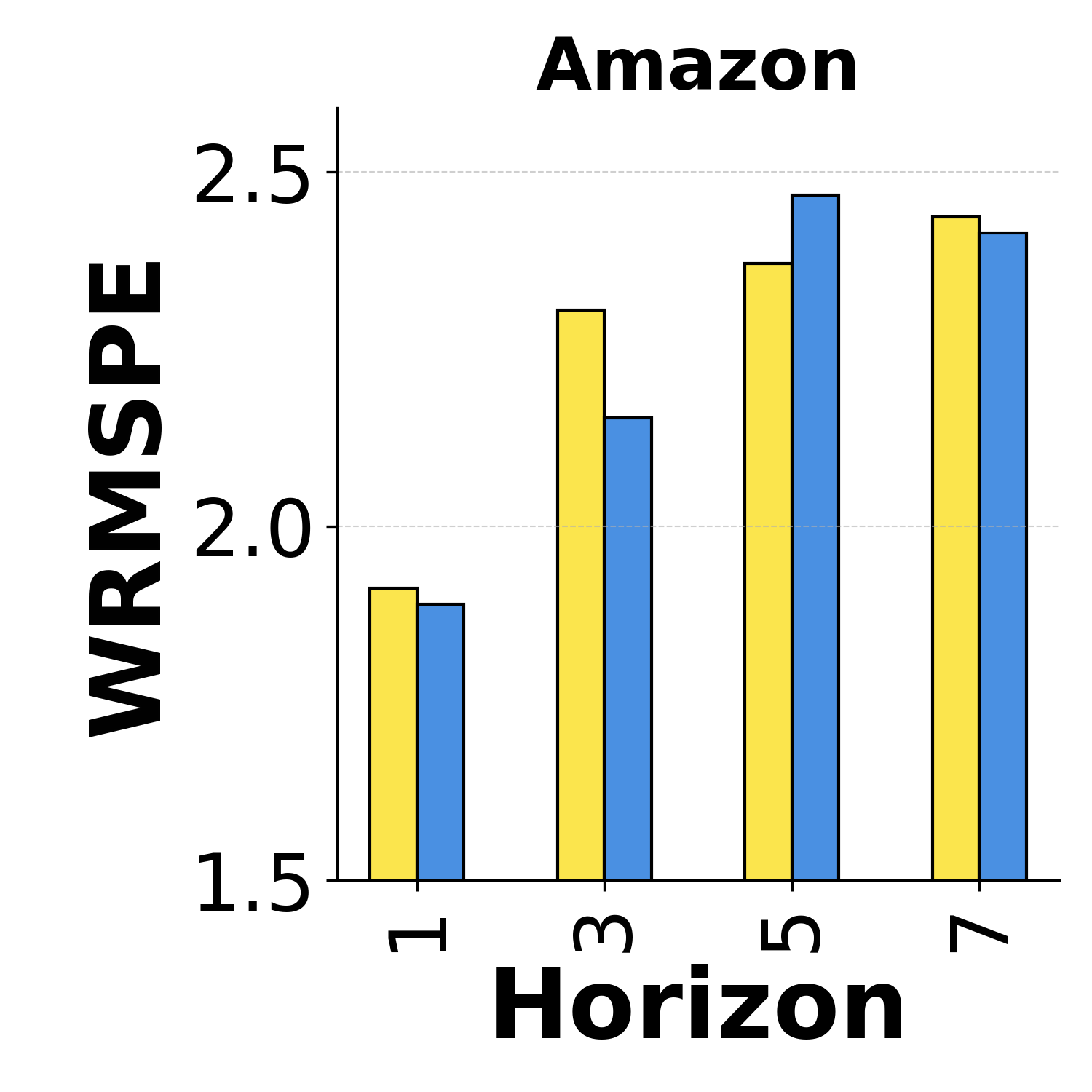}\hspace{2mm}
    \includegraphics[width=0.14\textwidth]{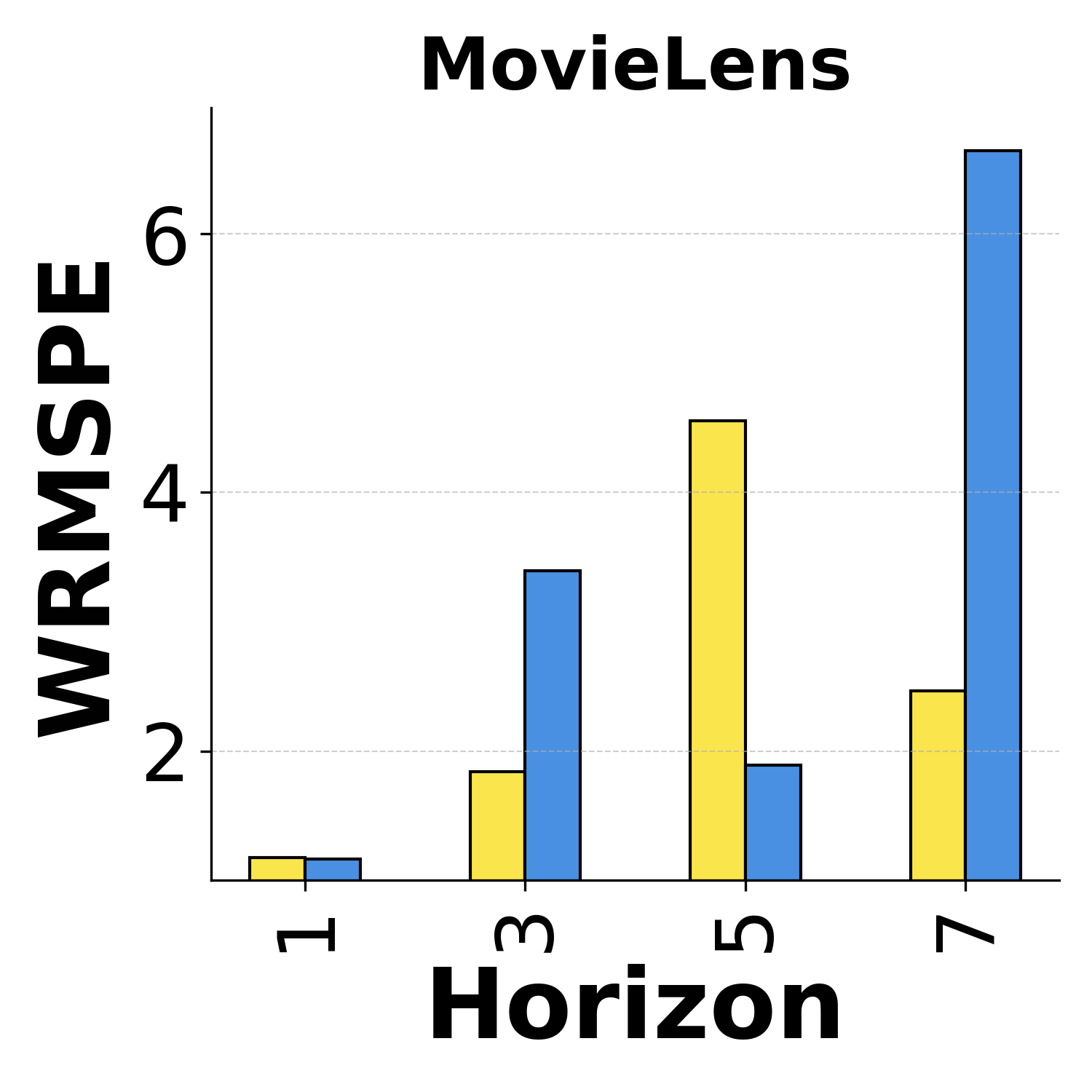}\hspace{2mm}
    \includegraphics[width=0.14\textwidth]{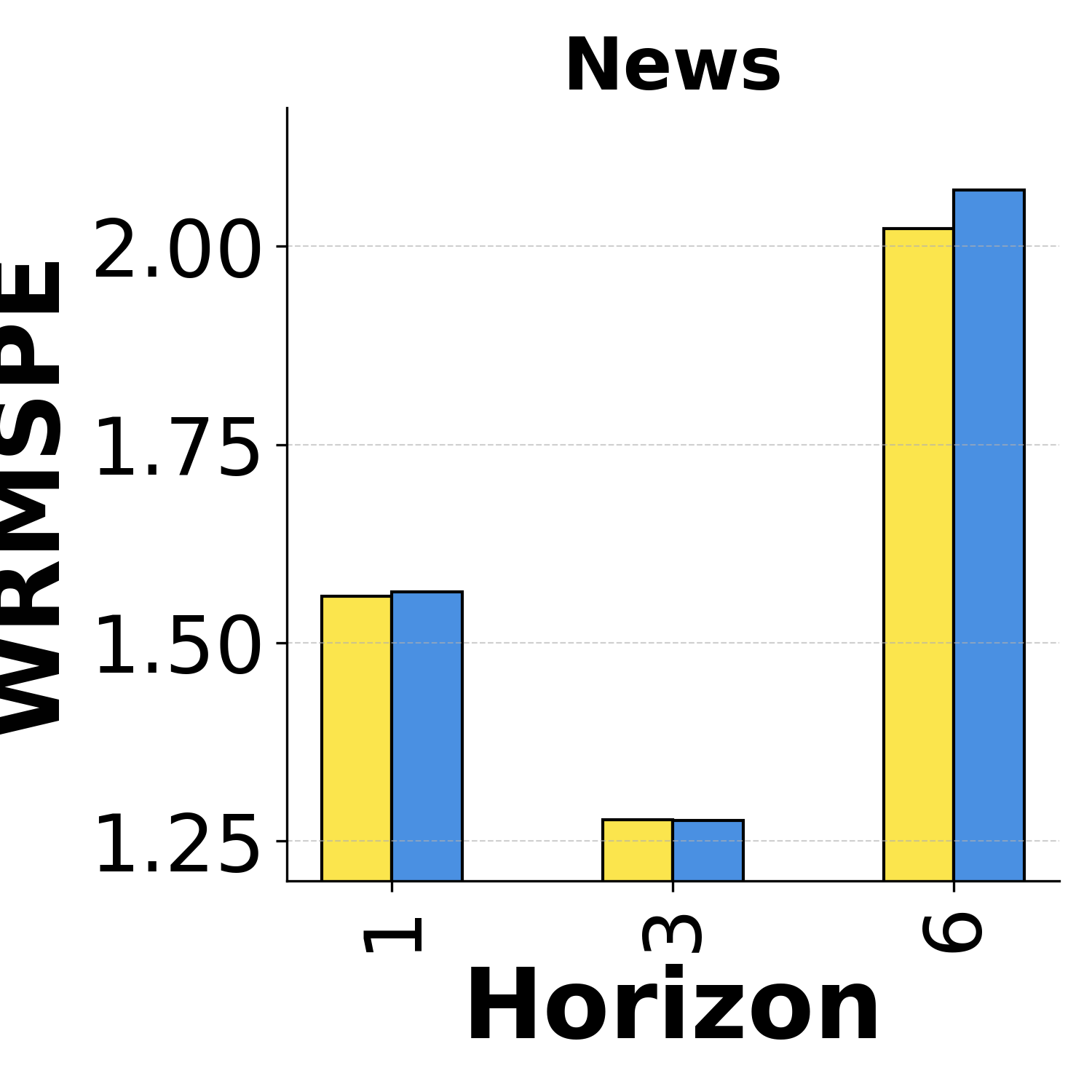}\hspace{2mm}
    \includegraphics[width=0.14\textwidth]{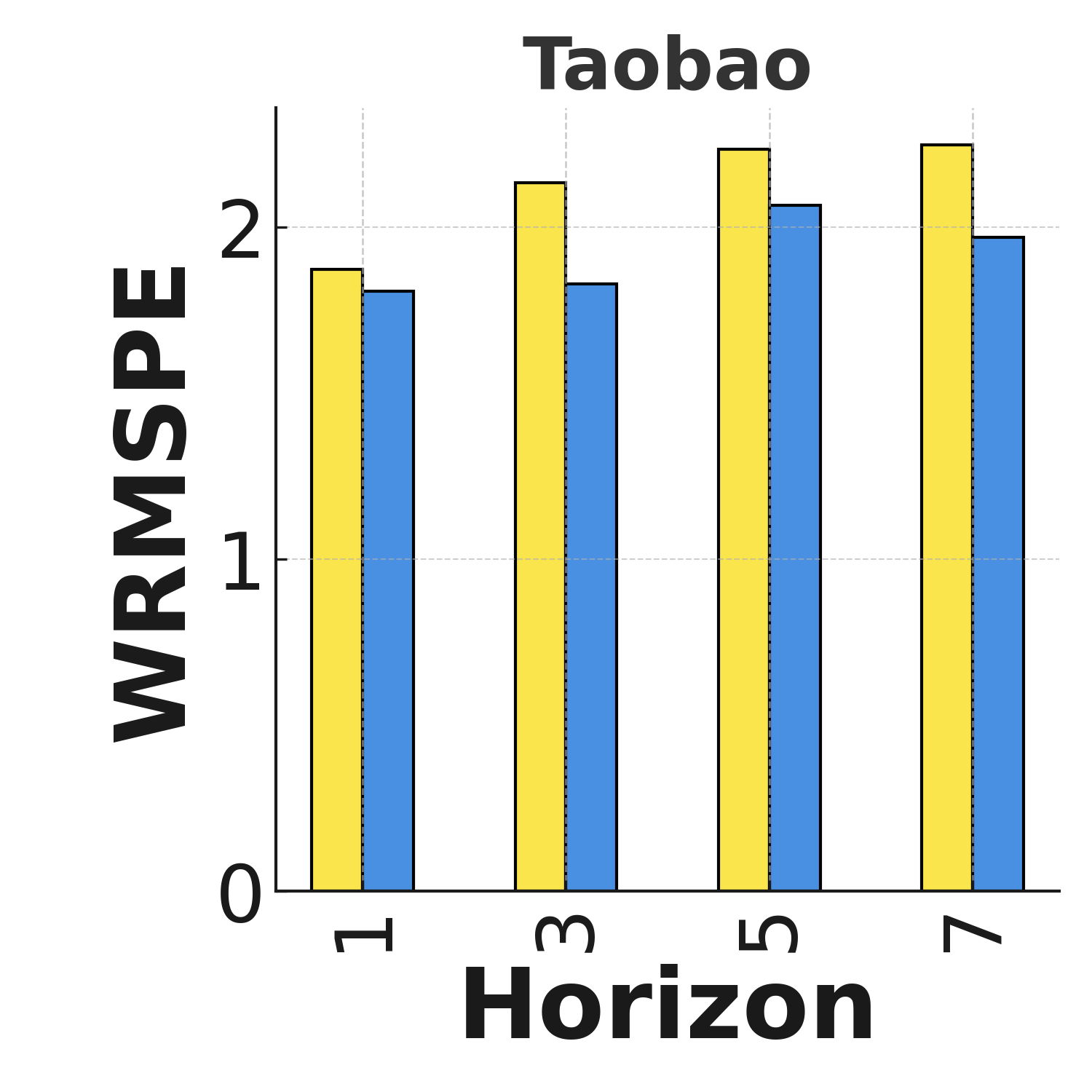}\hspace{2mm}
    \includegraphics[width=0.14\textwidth]{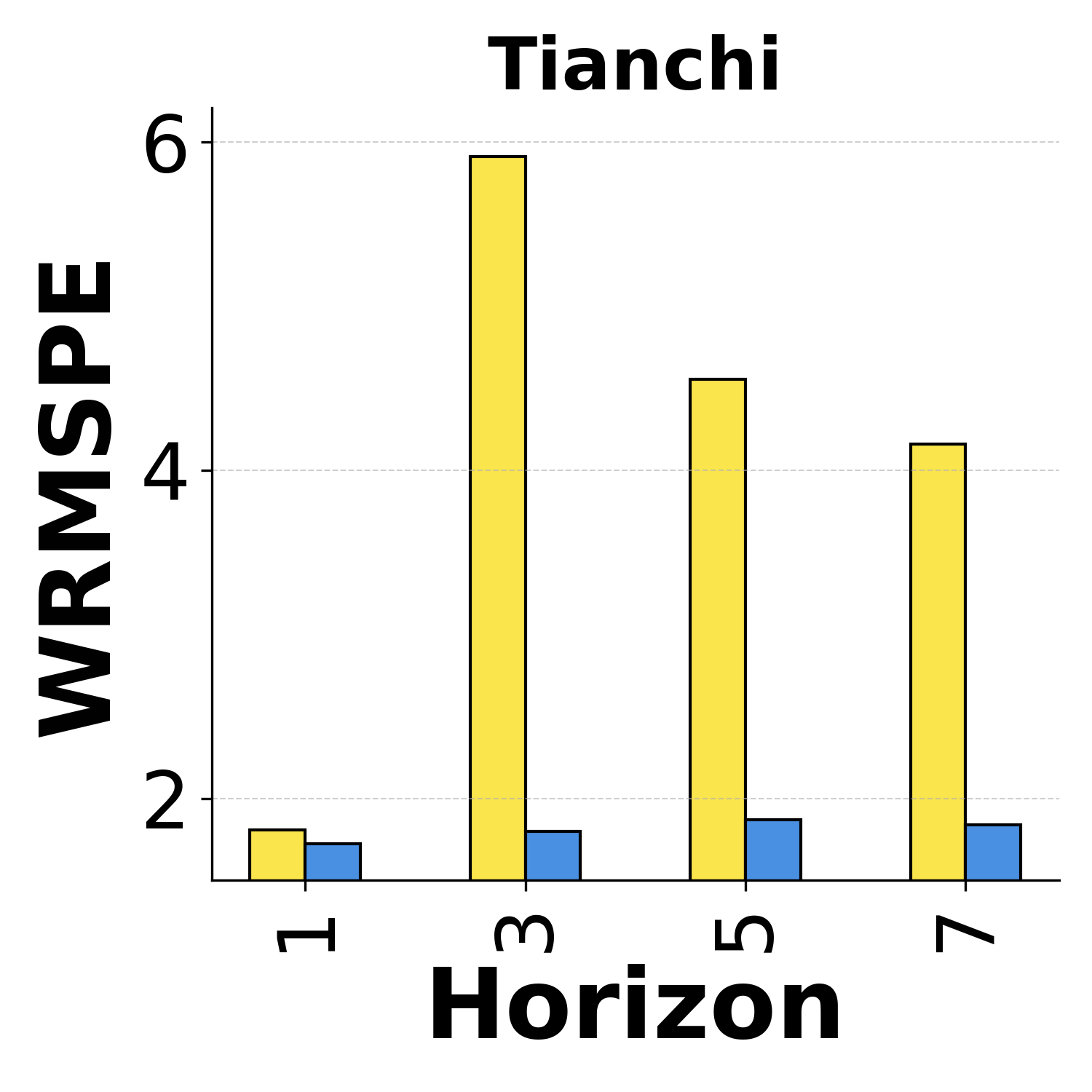}\hspace{2mm}
    \includegraphics[width=0.14\textwidth]{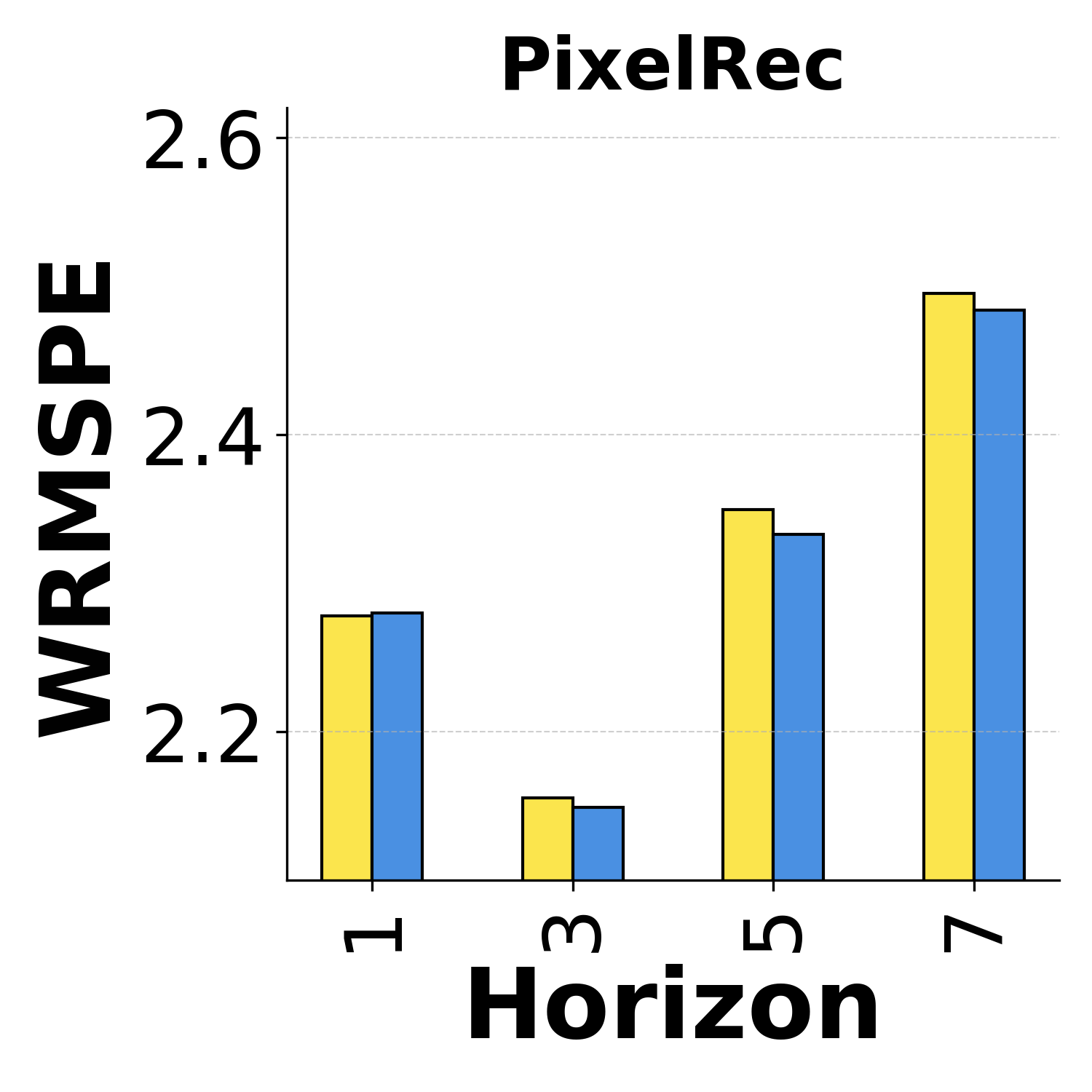}
    
    \caption{Cold-start forecasting results. Top: RMSE; Bottom: WRMSPE. The yellow bar is the baseline \texttt{average}, and the blue bar is \texttt{GPT} forecasting.}
    \label{fig:cold_start}
\end{figure}

Figure~\ref{fig:cold_start} summarizes the cold-start forecasting results across six datasets using both RMSE and WRMSPE. Several consistent patterns emerge:
Across most datasets, \texttt{GPT} generation consistently outperforms \texttt{average} baselines, verifying the effectiveness of modality-driven forecasting under extreme data sparsity. The improvement is particularly evident in sparse datasets like \texttt{PixelRec} and \texttt{Tianchi}, while \texttt{Movielens} shows a distinct advantage due to its spiky dense patterns, compared with sparse forecasting. Dataset-specific results and analyses are provided in Appendix~\ref{appendix:coldstart}.

\section{Discussion and Limitations}
\label{sec:disscussion}
While the results across datasets demonstrate the utility of external modalities in different scenarios, we also acknowledge several limitations in our current setup.

First, although \texttt{short} series naturally contain fewer windows than \texttt{long} series, we did not apply any upsampling strategy to balance them during training. As a result, short series may contribute less to gradient updates, potentially underestimating the real benefit of multimodalities in highly data-scarce cases. That said, we explicitly report results separately on \texttt{short} and \texttt{long} series, which still allows us to observe modality effects under different lengths. 
Besides, while we focus primarily on varying-history series, we recognize that sparsity is an equally important factor. A long series with mostly zeros or missing values may carry less signal than a shorter, denser one. This is especially relevant for behavioral data such as user interactions or item purchases. Our results suggest that external modalities are most helpful when the signal is weak, whether due to shortness or sparsity. However, a systematic investigation of sparsity effects is beyond the current scope, and we leave it to future work.
Finally, we note that using \texttt{GPT} for cold-start forecasting occasionally deviates from the target forecast horizon, despite explicit formatting in the prompt. While such cases are rare and do not significantly affect aggregate metrics, they reflect the inherent variability of large language models when used in non-autoregressive, multi-step prediction settings. In future work, stronger decoding constraints or horizon-aware prompting may improve consistency.

\section{Conclusion}
We present \textbf{MoTime}, a multimodal time series dataset suite designed to support structured evaluation under varying-history lengths and cold-start conditions for forecasting. It enables analysis of how external modalities can improve forecasting when time series data is limited.
\texttt{MoTime} provides a solid foundation for evaluating how different modalities help with forecasting. It also opens up future work on factors like data sparsity and how well relevant information can be retrieved.

\section{Appendix}

\subsection{Dataset Details}
\label{sec:dataset_details}
\subsubsection{PixelRec}
PixelRec~\cite{cheng2023image} is a multimodal dataset designed for research on recommender systems with content consumption, such as rich visual and behavioral signals, particularly in lifestyle, food, entertainment, and technology domains. It captures real-world user engagement with short videos and images, integrating high-level content signals such as visual thumbnails, titles, and tags, alongside user interaction metrics including view count, likes, comments, shares, and favorites. Each video entry includes detailed metadata such as release time, video duration, and trip-related tags, making it ideal for series-aware recommendation tasks.

To enable time series modeling, we transform the original user-item-timestamp triplets from PixelRec into a set of daily popularity series for items. Specifically, we aggregate user interaction signals on a per-item, per-day basis to construct time series reflecting item popularity dynamics over time. The resulting time series spans a total of 921 days from March 5, 2020, to September 12, 2022. However, the data is temporally sparse, with entries only on days when user interaction occurred, resulting in a non-continuous time series. The temporal granularity is at the daily level, and each item’s series length varies depending on its active lifespan and engagement frequency. To ensure statistical rigor, we remove incomplete days at both ends of the time span.

\subsubsection{TaobaoFashion}
The TaobaoFashion dataset originates from the Alibaba Tianchi ``Fashion Outfit Recommendation'' competition\footnote{\url{https://tianchi.aliyun.com/competition/entrance/231506/information}}. It is designed to support research in personalized fashion recommendation systems and includes multimodal item-level metadata such as product images, category information, and outfit combinations curated by stylists.

To construct a time series forecasting task, we process the anonymized user behavior logs to create daily-level purchase count series per item. These series reflect item-level popularity fluctuations across one year, from 2014-06-15 to 2015-06-14. The resulting temporal signals are sparse and non-continuous, mirroring real-world consumer behavior.

Each item in the dataset is linked to a static image, and we align these modalities via a unified item ID. The combination of image modality and purchase dynamics makes the dataset particularly suitable for short-horizon, image-conditioned popularity forecasting and demand trend analysis in online retail scenarios.

\subsubsection{MovieLens}
The MovieLens dataset\footnote{\url{https://grouplens.org/datasets/movielens/32m/}} is widely used in recommendation and trend analysis research. For our purposes, we transform this interaction dataset into a multimodal time series benchmark by aggregating daily-level user interactions (e.g., ratings) per movie.

Each time series reflects the cumulative popularity of a movie over time, spanning from 1995-01-09 to 2023-10-13. The series is typically long and sparse, depending on the movie's release date and user engagement.
External modalities are not directly included in the original dataset. To support multimodal modeling, we collected and aligned additional metadata per movie, including textual attributes such as title, release year, genres, user-generated tags, and overviews from public sources. Text fields such as ``overview'' and ``tags'' are cleaned, and missing or invalid entries are marked appropriately, e.g.-1 or NaN.
The genre coverage is diverse, including Action, Comedy, Documentary, Romance, Sci-Fi, and others, allowing the exploration of semantic-enhanced forecasting under sparse historical conditions. The inclusion of release year enables the simulation of cold-start settings by masking pre-release periods.

\subsubsection{AmazonReview}
The AmazonReview dataset is derived from the Amazon Reviews 2023 public release~\cite{hou2024bridging}, originally designed for product recommendation and sentiment analysis. For forecasting purposes, we convert the review log data into item-oriented daily time series by aggregating the number of reviews each item receives per day.

Each item series reflects review activity across 3,934 days, covering a time span from 2013-01-01 to 2023-10-01. The series are sparse and exhibit product lifecycle dynamics. We retain only items with at least 100 reviews to ensure sufficient series length and signal. Time series values take on three interpretations: -1 indicates the item had not yet been listed, 0 means the item was listed but received no reviews that day, and positive integers denote actual review counts. This encoding supports cold-start modeling, as early periods before product launch can be explicitly masked. External modalities, including product titles, descriptions, features, categories, prices, and ratings, are retained and aligned with each item using a unified identifier. The dataset includes 29 product categories, such as Books, Electronics, and Home, each representing a semantically distinct sub-dataset. Detailed category-level statistics are reported in Appendix~\ref{tab:amazon_datasets}, and the time series statistics are in~\ref{tab:amazon_stats}.

\begin{table}
    \centering
    \caption{Overview of the 29 AmazonReview Subsets}
    \label{tab:amazon_datasets}
    \begin{tabular}{lrr}
        \toprule
        Dataset & Channels (N) & Density (\%) \\
        \midrule
        All\_Beauty & 711 & 4.50 \\
        Amazon\_Fashion & 1060 & 5.00 \\
        Appliances & 3504 & 6.10 \\
        Arts\_Crafts\_and\_Sewing & 13748 & 5.81 \\
        Automotive & 27963 & 6.16 \\
        Baby\_Products & 8970 & 8.13 \\
        Beauty\_and\_Personal\_Care & 37528 & 7.23 \\
        Books & 25868 & 5.15 \\
        CDs\_and\_Vinyl & 3708 & 4.34 \\
        Cell\_Phones\_and\_Accessories & 31219 & 6.07 \\
        Clothing\_Shoes\_and\_Jewelry & 86574 & 7.16 \\
        Digital\_Music & 9 & 3.88 \\
        Electronics & 61121 & 7.11 \\
        Grocery\_and\_Gourmet\_Food & 23234 & 6.84 \\
        Handmade\_Products & 497 & 4.38 \\
        Health\_and\_Household & 38126 & 7.70 \\
        Health\_and\_Personal\_Care & 558 & 5.46 \\
        Home\_and\_Kitchen & 97172 & 7.20 \\
        Industrial\_and\_Scientific & 7596 & 6.33 \\
        Kindle\_Store & 40150 & 4.71 \\
        Movies\_and\_TV & 25428 & 4.95 \\
        Musical\_Instruments & 4589 & 6.42 \\
        Office\_Products & 20546 & 6.67 \\
        Patio\_Lawn\_and\_Garden & 27374 & 6.48 \\
        Pet\_Supplies & 23803 & 8.60 \\
        Sports\_and\_Outdoors & 28635 & 6.51 \\
        Tools\_and\_Home\_Improvement & 44069 & 6.55 \\
        Toys\_and\_Games & 28259 & 5.80 \\
        Video\_Games & 6737 & 6.91 \\
        \bottomrule
    \end{tabular}
\end{table}

\begin{table}
    \centering
    \caption{Per-category statistics of the Amazon Review dataset. The minimum value here refers to the smallest positive value that is neither zero nor -1.}
    \label{tab:amazon_stats}
    \begin{tabular}{lrrrr}
    \toprule
    \textbf{Category} & \textbf{Median} & \textbf{Mean} & \textbf{Min} & \textbf{Max} \\
    \midrule
    Industrial\_and\_Scientific & 1 & 1.19 & 1 & 257 \\
    Baby\_Products & 1 & 1.20 & 1 & 258 \\
    CDs\_and\_Vinyl & 1 & 1.19 & 1 & 194 \\
    Clothing\_Shoes\_and\_Jewelry & 1 & 1.22 & 1 & 159 \\
    Toys\_and\_Games & 1 & 1.20 & 1 & 1021 \\
    Video\_Games & 1 & 1.30 & 1 & 1370 \\
    Office\_Products & 1 & 1.19 & 1 & 714 \\
    Musical\_Instruments & 1 & 1.14 & 1 & 66 \\
    Home\_and\_Kitchen & 1 & 1.23 & 1 & 245 \\
    Kindle\_Store & 1 & 1.56 & 1 & 429 \\
    Tools\_and\_Home\_Improvement & 1 & 1.19 & 1 & 292 \\
    Health\_and\_Personal\_Care & 1 & 1.22 & 1 & 82 \\
    Beauty\_and\_Personal\_Care & 1 & 1.23 & 1 & 524 \\
    Patio\_Lawn\_and\_Garden & 1 & 1.20 & 1 & 130 \\
    Arts\_Crafts\_and\_Sewing & 1 & 1.16 & 1 & 139 \\
    All\_Beauty & 1 & 1.20 & 1 & 19 \\
    Health\_and\_Household & 1 & 1.26 & 1 & 630 \\
    Sports\_and\_Outdoors & 1 & 1.18 & 1 & 596 \\
    Automotive & 1 & 1.14 & 1 & 75 \\
    Movies\_and\_TV & 1 & 1.73 & 1 & 6311 \\
    Books & 1 & 1.32 & 1 & 709 \\
    Handmade\_Products & 1 & 1.14 & 1 & 22 \\
    Pet\_Supplies & 1 & 1.24 & 1 & 394 \\
    Electronics & 1 & 1.32 & 1 & 4248 \\
    Grocery\_and\_Gourmet\_Food & 1 & 1.16 & 1 & 212 \\
    Digital\_Music & 1 & 1.08 & 1 & 4 \\
    Appliances & 1 & 1.19 & 1 & 200 \\
    Cell\_Phones\_and\_Accessories & 1 & 1.36 & 1 & 223 \\
    Amazon\_Fashion & 1 & 1.20 & 1 & 1000 \\
    \bottomrule
    \end{tabular}
\end{table}

\subsubsection{Tianchi}
The Tianchi dataset is released by Alibaba Group via the Tianchi platform\footnote{\url{https://tianchi.aliyun.com/dataset/43}} and is designed for large-scale behavior-based recommendation research. It includes extensive user-item interaction logs, item metadata, and user-generated content such as reviews.

To construct a forecasting task, we aggregate daily purchase counts for each item, transforming interaction logs into item-oriented time series. The series spans a full year and captures realistic purchasing trends, including bursty behavior and long-tail activity. Each item is associated with multiple static attributes, including textual product titles, which are processed as keyword series, hierarchical category paths, and image URLs. We align time series, text, and image modalities using a unified item ID within the dataset. The multimodal structure of the dataset, combined with its high item volume and behavioral richness, supports a range of tasks, such as cold-start prediction, trend analysis, and modality-ablation benchmarking under real-world e-commerce conditions.

\subsubsection{News}
The News dataset is based on the popularity prediction benchmark introduced by Moniz and Torgo~\cite{moniz2018news}, which tracks early-stage user engagement with news articles across multiple social media platforms. It provides high-resolution temporal signals with multimodal metadata.

Each news article is associated with a time series of 144 values, capturing popularity over 20-minute intervals for the first 48 hours after publication. These series reflect how quickly an article gains traction and fades in user attention, making the dataset ideal for fine-grained popularity forecasting.
External modalities include the article’s title, headline, publishing source, topic tag, e.g., ``obama'', ``economy'', timestamp, and sentiment polarity of the title and headline. These features allow joint modeling of textual semantics and temporal dynamics.
In our benchmark, we focus on a filtered subset of the data where the topic is ``obama'' and the platform is Facebook. This setting captures attention dynamics in a real-world political news domain, with well-defined short-term temporal patterns and content relevance.

\subsubsection{WikiPeople}
The WikiPeople dataset is derived from the Wikipedia web traffic dataset~\cite{yang2017web} originally released in the Kaggle Web Traffic Time Series Forecasting competition. This curated subset focuses specifically on Wikipedia articles about people, such as public figures, artists, scientists, and politicians.

Each article is associated with daily page view counts across multiple access channels, including desktop, mobile web, and all-access. These multichannel time series offer insights into how different devices contribute to content consumption over time, introducing cross-channel variation.
To enrich the data with semantic context, we collect short English-language summaries for a subset of 918 articles by scraping Wikipedia content. These summaries are aligned to the time series via article IDs and serve as an external textual modality.
The dataset enables a variety of tasks, such as channel-specific forecasting, device trend analysis, and text-conditioned prediction of popularity fluctuations. Its real-world heterogeneity and multichannel structure make it a strong benchmark for testing multimodal series modeling methods.

\subsection{Statistics of Time Series}
The statistics of time series values are shown in Table~\ref{tab:ts_stats}.

\begin{table}
    \centering
    \caption{Descriptive statistics of time series values per dataset. The minimum value here refers to the smallest positive value that is neither zero nor -1.}
    \label{tab:ts_stats}
    \
    \begin{tabular}{lrrrr}
        \toprule
                 \textbf{Dataset} & \textbf{Median} & \textbf{Mean} & \textbf{Min} & \textbf{Max} \\
        \midrule
        PixelRec & 2 & 13.87 & 1 & 3196 \\
        Tianchi & 7 & 46.74 & 1 & 90472 \\
        MovieLens & 1 & 1.08 & 1 & 549 \\
        News & 1 & 11.63 & 1 & 13291 \\
        TaobaoFashion & 3 & 5.36 & 1 & 966 \\
        WikiPeople & 921.25 & 3165.67 & 1 & 5816910 \\
        AmazonReview & 1 & 1.24 & 1 & 6311 \\
        \bottomrule
    \end{tabular}
\end{table}

\subsection{Statistics of Text}
We statistically analyze the textual modality from different granularities. Table~\ref{tab:character_length_stats}, Table~\ref{tab:word_length_stats}, Table~\ref{tab:sentence_length_stats}, Table~\ref{tab:llama_token_length_stats}, Table~\ref{tab:gpt_token_length_stats} show the number of characters, words, sentences, tokens in Llama and GPT standard, separately.

\begin{table}[htbp]
\centering
\caption{Character Length Statistics per Dataset.}
\resizebox{\textwidth}{!}{
\begin{tabular}{l|r|r|r|r}
\toprule
                                           dataset &  char\_len\_min &  char\_len\_max &  char\_len\_mean &  char\_len\_median \\
\midrule
                                  PixelRec &            85 &          2458 &         330.60 &              260 \\
                                           Tianchi &             1 &            94 &          47.41 &               48 \\
                                         MovieLens &           149 &         85368 &         738.19 &              479 \\
                                           News &           261 &           786 &         460.09 &              438 \\
                    Books &          3718 &          3718 &        3718.00 &             3718 \\
            Movies\_and\_TV &          3611 &          3611 &        3611.00 &             3611 \\
              Video\_Games &          3789 &          3789 &        3789.00 &             3789 \\
 Grocery\_and\_Gourmet\_Food &          3895 &          3895 &        3895.00 &             3895 \\
   Arts\_Crafts\_and\_Sewing &          3894 &          3894 &        3894.00 &             3894 \\
      Sports\_and\_Outdoors &          3840 &          3840 &        3840.00 &             3840 \\
             Pet\_Supplies &          3796 &          3796 &        3796.00 &             3796 \\
Clothing\_Shoes\_and\_Jewelry &          3817 &          3817 &        3817.00 &             3817 \\
               Appliances &          3921 &          3921 &        3921.00 &             3921 \\
      Musical\_Instruments &          3877 &          3877 &        3877.00 &             3877 \\
        Handmade\_Products &          3738 &          3738 &        3738.00 &             3738 \\
               All\_Beauty &          3793 &          3793 &        3793.00 &             3793 \\
 Beauty\_and\_Personal\_Care &          3906 &          3906 &        3906.00 &             3906 \\
          Office\_Products &          3938 &          3938 &        3938.00 &             3938 \\
            Digital\_Music &          2714 &          2714 &        2714.00 &             2714 \\
            Baby\_Products &          3899 &          3899 &        3899.00 &             3899 \\
           Amazon\_Fashion &          3800 &          3800 &        3800.00 &             3800 \\
Tools\_and\_Home\_Improvement &          3917 &          3917 &        3917.00 &             3917 \\
 Health\_and\_Personal\_Care &          3903 &          3903 &        3903.00 &             3903 \\
    Patio\_Lawn\_and\_Garden &          3950 &          3950 &        3950.00 &             3950 \\
               Automotive &          3785 &          3785 &        3785.00 &             3785 \\
           Toys\_and\_Games &          3906 &          3906 &        3906.00 &             3906 \\
              Electronics &          3938 &          3938 &        3938.00 &             3938 \\
Industrial\_and\_Scientific &          3920 &          3920 &        3920.00 &             3920 \\
             Kindle\_Store &          3773 &          3773 &        3773.00 &             3773 \\
         Home\_and\_Kitchen &          3928 &          3928 &        3928.00 &             3928 \\
Cell\_Phones\_and\_Accessories &          3938 &          3938 &        3938.00 &             3938 \\
            CDs\_and\_Vinyl &          3788 &          3788 &        3788.00 &             3788 \\
     Health\_and\_Household &          3905 &          3905 &        3905.00 &             3905 \\
\bottomrule
\end{tabular}}
\label{tab:character_length_stats}
\end{table}

\begin{table}[htbp]
\centering
\caption{Word Length Statistics per Dataset.}
\resizebox{0.9\textwidth}{!}{
\begin{tabular}{l|r|r|r|r}
\toprule
                                           dataset &  word\_len\_min &  word\_len\_max &  word\_len\_mean &  word\_len\_median \\
\midrule
                                  PixelRec &            17 &           430 &          66.99 &               53 \\
                                           Tianchi &             1 &            29 &          12.86 &               13 \\
                                         MovieLens &            33 &          5302 &         105.58 &               86 \\
                                           News &            54 &           163 &          88.38 &               84 \\
                    Books &           575 &           575 &         575.00 &              575 \\
            Movies\_and\_TV &           439 &           439 &         439.00 &              439 \\
              Video\_Games &           498 &           498 &         498.00 &              498 \\
 Grocery\_and\_Gourmet\_Food &           549 &           549 &         549.00 &              549 \\
   Arts\_Crafts\_and\_Sewing &           575 &           575 &         575.00 &              575 \\
      Sports\_and\_Outdoors &           519 &           519 &         519.00 &              519 \\
             Pet\_Supplies &           550 &           550 &         550.00 &              550 \\
Clothing\_Shoes\_and\_Jewelry &           539 &           539 &         539.00 &              539 \\
               Appliances &           548 &           548 &         548.00 &              548 \\
      Musical\_Instruments &           509 &           509 &         509.00 &              509 \\
        Handmade\_Products &           534 &           534 &         534.00 &              534 \\
               All\_Beauty &           468 &           468 &         468.00 &              468 \\
 Beauty\_and\_Personal\_Care &           575 &           575 &         575.00 &              575 \\
          Office\_Products &           541 &           541 &         541.00 &              541 \\
            Digital\_Music &           375 &           375 &         375.00 &              375 \\
            Baby\_Products &           556 &           556 &         556.00 &              556 \\
           Amazon\_Fashion &           503 &           503 &         503.00 &              503 \\
Tools\_and\_Home\_Improvement &           529 &           529 &         529.00 &              529 \\
 Health\_and\_Personal\_Care &           540 &           540 &         540.00 &              540 \\
    Patio\_Lawn\_and\_Garden &           527 &           527 &         527.00 &              527 \\
               Automotive &           512 &           512 &         512.00 &              512 \\
           Toys\_and\_Games &           568 &           568 &         568.00 &              568 \\
              Electronics &           518 &           518 &         518.00 &              518 \\
Industrial\_and\_Scientific &           550 &           550 &         550.00 &              550 \\
             Kindle\_Store &           562 &           562 &         562.00 &              562 \\
         Home\_and\_Kitchen &           542 &           542 &         542.00 &              542 \\
Cell\_Phones\_and\_Accessories &           533 &           533 &         533.00 &              533 \\
            CDs\_and\_Vinyl &           454 &           454 &         454.00 &              454 \\
     Health\_and\_Household &           533 &           533 &         533.00 &              533 \\
\bottomrule
\end{tabular}}
\label{tab:word_length_stats}
\end{table}

\begin{table}[htbp]
\centering
\caption{Sentence Length Statistics per Dataset}
    \resizebox{0.8\textwidth}{!}{
\begin{tabular}{l|r|r|r|r}
\toprule
                                           dataset &  sent\_len\_min &  sent\_len\_max &  sent\_len\_mean &  sent\_len\_median \\
\midrule
                                  PixelRec &             2 &            86 &           3.53 &                3 \\
                                   PixelRec/subset &             3 &            15 &           4.66 &                4 \\
                                           Tianchi &             1 &             3 &           1.01 &                1 \\
                                         MovieLens &             3 &            62 &           5.44 &                5 \\
                                           News &             4 &            13 &           6.47 &                6 \\
                    Books &            12 &            12 &          12.00 &               12 \\
            Movies\_and\_TV &             8 &             8 &           8.00 &                8 \\
              Video\_Games &             9 &             9 &           9.00 &                9 \\
 Grocery\_and\_Gourmet\_Food &            10 &            10 &          10.00 &               10 \\
   Arts\_Crafts\_and\_Sewing &             9 &             9 &           9.00 &                9 \\
      Sports\_and\_Outdoors &             8 &             8 &           8.00 &                8 \\
             Pet\_Supplies &            11 &            11 &          11.00 &               11 \\
Clothing\_Shoes\_and\_Jewelry &             8 &             8 &           8.00 &                8 \\
               Appliances &             8 &             8 &           8.00 &                8 \\
      Musical\_Instruments &             8 &             8 &           8.00 &                8 \\
        Handmade\_Products &            11 &            11 &          11.00 &               11 \\
               All\_Beauty &             9 &             9 &           9.00 &                9 \\
 Beauty\_and\_Personal\_Care &            12 &            12 &          12.00 &               12 \\
          Office\_Products &             9 &             9 &           9.00 &                9 \\
            Digital\_Music &             8 &             8 &           8.00 &                8 \\
            Baby\_Products &             8 &             8 &           8.00 &                8 \\
           Amazon\_Fashion &             8 &             8 &           8.00 &                8 \\
Tools\_and\_Home\_Improvement &             9 &             9 &           9.00 &                9 \\
 Health\_and\_Personal\_Care &            10 &            10 &          10.00 &               10 \\
    Patio\_Lawn\_and\_Garden &            14 &            14 &          14.00 &               14 \\
               Automotive &             9 &             9 &           9.00 &                9 \\
           Toys\_and\_Games &            10 &            10 &          10.00 &               10 \\
              Electronics &             8 &             8 &           8.00 &                8 \\
Industrial\_and\_Scientific &             8 &             8 &           8.00 &                8 \\
             Kindle\_Store &            10 &            10 &          10.00 &               10 \\
         Home\_and\_Kitchen &             9 &             9 &           9.00 &                9 \\
Cell\_Phones\_and\_Accessories &             8 &             8 &           8.00 &                8 \\
            CDs\_and\_Vinyl &             9 &             9 &           9.00 &                9 \\
     Health\_and\_Household &             8 &             8 &           8.00 &                8 \\
\bottomrule
\end{tabular}}
\label{tab:sentence_length_stats}
\end{table}

\begin{table}[htbp]
\centering
\caption{GPT Token Length Statistics per Dataset}
    \resizebox{\textwidth}{!}{
\begin{tabular}{l|r|r|r|r}
\toprule
                                           dataset &  token\_len\_gpt\_min &  token\_len\_gpt\_max &  token\_len\_gpt\_mean &  token\_len\_gpt\_median \\
\midrule
                                  PixelRec &                 17 &                462 &               73.33 &                    58 \\
                                   PixelRec/subset &                 46 &                300 &               99.23 &                    84 \\
                                           Tianchi &                  1 &                 75 &               50.15 &                    51 \\
                                         MovieLens &                 35 &              23554 &              182.49 &                   108 \\
                                           News &                 66 &                187 &              103.89 &                   100 \\
                    Books &               1049 &               1049 &             1049.00 &                  1049 \\
            Movies\_and\_TV &               1157 &               1157 &             1157.00 &                  1157 \\
              Video\_Games &               1165 &               1165 &             1165.00 &                  1165 \\
 Grocery\_and\_Gourmet\_Food &               1268 &               1268 &             1268.00 &                  1268 \\
   Arts\_Crafts\_and\_Sewing &               1280 &               1280 &             1280.00 &                  1280 \\
      Sports\_and\_Outdoors &               1217 &               1217 &             1217.00 &                  1217 \\
             Pet\_Supplies &               1313 &               1313 &             1313.00 &                  1313 \\
Clothing\_Shoes\_and\_Jewelry &               1325 &               1325 &             1325.00 &                  1325 \\
               Appliances &               1304 &               1304 &             1304.00 &                  1304 \\
      Musical\_Instruments &               1244 &               1244 &             1244.00 &                  1244 \\
        Handmade\_Products &               1272 &               1272 &             1272.00 &                  1272 \\
               All\_Beauty &               1129 &               1129 &             1129.00 &                  1129 \\
 Beauty\_and\_Personal\_Care &               1298 &               1298 &             1298.00 &                  1298 \\
          Office\_Products &               1299 &               1299 &             1299.00 &                  1299 \\
            Digital\_Music &                905 &                905 &              905.00 &                   905 \\
            Baby\_Products &               1295 &               1295 &             1295.00 &                  1295 \\
           Amazon\_Fashion &               1215 &               1215 &             1215.00 &                  1215 \\
Tools\_and\_Home\_Improvement &               1273 &               1273 &             1273.00 &                  1273 \\
 Health\_and\_Personal\_Care &               1248 &               1248 &             1248.00 &                  1248 \\
    Patio\_Lawn\_and\_Garden &               1275 &               1275 &             1275.00 &                  1275 \\
               Automotive &               1246 &               1246 &             1246.00 &                  1246 \\
           Toys\_and\_Games &               1289 &               1289 &             1289.00 &                  1289 \\
              Electronics &               1248 &               1248 &             1248.00 &                  1248 \\
Industrial\_and\_Scientific &               1358 &               1358 &             1358.00 &                  1358 \\
             Kindle\_Store &               1265 &               1265 &             1265.00 &                  1265 \\
         Home\_and\_Kitchen &               1248 &               1248 &             1248.00 &                  1248 \\
Cell\_Phones\_and\_Accessories &               1267 &               1267 &             1267.00 &                  1267 \\
            CDs\_and\_Vinyl &               1093 &               1093 &             1093.00 &                  1093 \\
     Health\_and\_Household &               1280 &               1280 &             1280.00 &                  1280 \\
\bottomrule
\end{tabular}}
\label{tab:gpt_token_length_stats}
\end{table}

\begin{table}[htbp]
\centering
\caption{BERT Token Length Statistics per Dataset}
    \resizebox{\textwidth}{!}{
\begin{tabular}{l|r|r|r|r}
\toprule
                                           dataset &  token\_len\_bert\_min &  token\_len\_bert\_max &  token\_len\_bert\_mean &  token\_len\_bert\_median \\
\midrule
                                  PixelRec &                  19 &                 777 &                77.29 &                     61 \\
                                   PixelRec/subset &                  49 &                 307 &               105.81 &                     89 \\
                                           Tianchi &                   3 &                  45 &                28.78 &                     30 \\
                                         MovieLens &                  35 &               21220 &               171.95 &                    107 \\
                                           News &                  66 &                 202 &               104.91 &                    101 \\
                    Books &                1075 &                1075 &              1075.00 &                   1075 \\
            Movies\_and\_TV &                1071 &                1071 &              1071.00 &                   1071 \\
              Video\_Games &                1141 &                1141 &              1141.00 &                   1141 \\
 Grocery\_and\_Gourmet\_Food &                1264 &                1264 &              1264.00 &                   1264 \\
   Arts\_Crafts\_and\_Sewing &                1280 &                1280 &              1280.00 &                   1280 \\
      Sports\_and\_Outdoors &                1181 &                1181 &              1181.00 &                   1181 \\
             Pet\_Supplies &                1301 &                1301 &              1301.00 &                   1301 \\
Clothing\_Shoes\_and\_Jewelry &                1295 &                1295 &              1295.00 &                   1295 \\
               Appliances &                1285 &                1285 &              1285.00 &                   1285 \\
      Musical\_Instruments &                1211 &                1211 &              1211.00 &                   1211 \\
        Handmade\_Products &                1256 &                1256 &              1256.00 &                   1256 \\
               All\_Beauty &                1120 &                1120 &              1120.00 &                   1120 \\
 Beauty\_and\_Personal\_Care &                1275 &                1275 &              1275.00 &                   1275 \\
          Office\_Products &                1231 &                1231 &              1231.00 &                   1231 \\
            Digital\_Music &                 864 &                 864 &               864.00 &                    864 \\
            Baby\_Products &                1290 &                1290 &              1290.00 &                   1290 \\
           Amazon\_Fashion &                1184 &                1184 &              1184.00 &                   1184 \\
Tools\_and\_Home\_Improvement &                1278 &                1278 &              1278.00 &                   1278 \\
 Health\_and\_Personal\_Care &                1238 &                1238 &              1238.00 &                   1238 \\
    Patio\_Lawn\_and\_Garden &                1257 &                1257 &              1257.00 &                   1257 \\
               Automotive &                1288 &                1288 &              1288.00 &                   1288 \\
           Toys\_and\_Games &                1264 &                1264 &              1264.00 &                   1264 \\
              Electronics &                1248 &                1248 &              1248.00 &                   1248 \\
Industrial\_and\_Scientific &                1323 &                1323 &              1323.00 &                   1323 \\
             Kindle\_Store &                1259 &                1259 &              1259.00 &                   1259 \\
         Home\_and\_Kitchen &                1246 &                1246 &              1246.00 &                   1246 \\
Cell\_Phones\_and\_Accessories &                1236 &                1236 &              1236.00 &                   1236 \\
            CDs\_and\_Vinyl &                1018 &                1018 &              1018.00 &                   1018 \\
     Health\_and\_Household &                1254 &                1254 &              1254.00 &                   1254 \\
\bottomrule
\end{tabular}}
\label{tab:bert_token_length_stats}
\end{table}

\begin{table}[htbp]
\centering
\caption{LLaMA Token Length Statistics per Dataset}
    \resizebox{\textwidth}{!}{
\begin{tabular}{l|r|r|r|r}
\toprule
                                           dataset &  token\_len\_llama\_min &  token\_len\_llama\_max &  token\_len\_llama\_mean &  token\_len\_llama\_median \\
\midrule
                                  PixelRec &                   20 &                  467 &                 85.16 &                      67 \\
                                   PixelRec/subset &                   54 &                  376 &                115.22 &                      98 \\
                                           Tianchi &                    3 &                   96 &                 63.70 &                      65 \\
                                         MovieLens &                   40 &                28119 &                216.53 &                     128 \\
                                           News &                   84 &                  231 &                130.75 &                     126 \\
                    Books &                 1633 &                 1633 &               1633.00 &                    1633 \\
            Movies\_and\_TV &                 1421 &                 1421 &               1421.00 &                    1421 \\
              Video\_Games &                 1516 &                 1516 &               1516.00 &                    1516 \\
 Grocery\_and\_Gourmet\_Food &                 1590 &                 1590 &               1590.00 &                    1590 \\
   Arts\_Crafts\_and\_Sewing &                 1658 &                 1658 &               1658.00 &                    1658 \\
      Sports\_and\_Outdoors &                 1583 &                 1583 &               1583.00 &                    1583 \\
             Pet\_Supplies &                 1648 &                 1648 &               1648.00 &                    1648 \\
Clothing\_Shoes\_and\_Jewelry &                 1646 &                 1646 &               1646.00 &                    1646 \\
               Appliances &                 1667 &                 1667 &               1667.00 &                    1667 \\
      Musical\_Instruments &                 1583 &                 1583 &               1583.00 &                    1583 \\
        Handmade\_Products &                 1591 &                 1591 &               1591.00 &                    1591 \\
               All\_Beauty &                 1507 &                 1507 &               1507.00 &                    1507 \\
 Beauty\_and\_Personal\_Care &                 1629 &                 1629 &               1629.00 &                    1629 \\
          Office\_Products &                 1591 &                 1591 &               1591.00 &                    1591 \\
            Digital\_Music &                 1187 &                 1187 &               1187.00 &                    1187 \\
            Baby\_Products &                 1642 &                 1642 &               1642.00 &                    1642 \\
           Amazon\_Fashion &                 1558 &                 1558 &               1558.00 &                    1558 \\
Tools\_and\_Home\_Improvement &                 1627 &                 1627 &               1627.00 &                    1627 \\
 Health\_and\_Personal\_Care &                 1578 &                 1578 &               1578.00 &                    1578 \\
    Patio\_Lawn\_and\_Garden &                 1629 &                 1629 &               1629.00 &                    1629 \\
               Automotive &                 1632 &                 1632 &               1632.00 &                    1632 \\
           Toys\_and\_Games &                 1603 &                 1603 &               1603.00 &                    1603 \\
              Electronics &                 1623 &                 1623 &               1623.00 &                    1623 \\
Industrial\_and\_Scientific &                 1693 &                 1693 &               1693.00 &                    1693 \\
             Kindle\_Store &                 1577 &                 1577 &               1577.00 &                    1577 \\
         Home\_and\_Kitchen &                 1623 &                 1623 &               1623.00 &                    1623 \\
Cell\_Phones\_and\_Accessories &                 1587 &                 1587 &               1587.00 &                    1587 \\
            CDs\_and\_Vinyl &                 1428 &                 1428 &               1428.00 &                    1428 \\
    Health\_and\_Household &                 1589 &                 1589 &               1589.00 &                    1589 \\
\bottomrule
\end{tabular}}
\label{tab:llama_token_length_stats}
\end{table}

\subsection{Unimodal Time Series Dataset and Benchmarks}
Unimodal forecasting resources primarily focus on numerical time series as the sole input modality. We group these resources into three categories: benchmarks with standardized evaluation protocols, dataset repositories that serve as foundations for model training and testing, and commonly used datasets in empirical studies.

\paragraph{Benchmark Suites.}
Recent benchmarks differ significantly in their design goals, evaluation protocols, and model compatibility. Among them, GIFT-Eval~\cite{woo2024gift}, Nixtla~\cite{nixtla2024benchmark}, and TSLib~\cite{zhang2024deep} are widely used for evaluating time series forecasting models. 
GIFT-Eval is tailored for foundation models—such as Moirai~\cite{wang2024moirai}, LagLLama~\cite{rasul2024lagllama}, TimesFM~\cite{das2024decoder}, Chronos~\cite{chronos2024}, and Granite~\cite{ekambaram2024ttm}—with a focus on rigorous zero-shot and fine-tuning protocols. However, it lacks support for incorporating entity-level context or external information.  
Nixtla emphasizes inference efficiency and deployment readiness, benchmarking models like TimeGPT-1~\cite{garza2025timegpt}, TimesFM, and Chronos. Yet, it provides limited insight into model behavior under semantic variability or data heterogeneity.  
TSLib provides a unified codebase for deep time series models across multiple tasks but does not define standardized forecasting protocols, making cross-model comparison less systematic.

\paragraph{Dataset Repositories.}
Several repositories offer large-scale time series data for model training and generalization evaluation. The Monash Time Series Forecasting Repository~\cite{godahewa2021monash} is widely used, with over 30 curated datasets from domains such as energy, healthcare, and finance.  
The UCI Machine Learning Repository~\cite{dua2017uci} includes a few datasets with temporal structure, occasionally repurposed for forecasting tasks in small-scale or adaptability-oriented studies.  
More recently, repositories such as LOTSA~\cite{wang2024moirai}, the Time-series Pile~\cite{zhang2024moment}, and Datadog Observability Metrics~\cite{cohen2024toto} have been introduced to support large-scale foundation model pretraining, containing up to hundreds of billions of time points. These collections are primarily designed for scale, rather than structured evaluation.

\paragraph{Commonly Used Datasets.}
A number of datasets have become de facto empirical testbeds in academic research, despite lacking formal benchmarking protocols. Wikipedia Page Views~\cite{wikikaggle2017,zhou2024scalable} is commonly used to evaluate model robustness under irregular and volatile dynamics. The M-competitions~\cite{makridakis2000m3,makridakis2018m4,makridakis2022m5} provide univariate time series from various industries and frequencies, and remain influential in evaluating forecasting performance.  
In addition, datasets from the Informer benchmark~\cite{zhou2021informer}, including Electricity, Traffic, Weather, Exchange, ETTh, and ETTm, are widely adopted for assessing long-horizon forecasting across domains such as energy, transportation, and industrial systems. However, these datasets are static in scope and not designed to probe generalization capabilities.

While benchmark, repositories, and datasets have collectively advanced time series forecasting, they all share several limitations: they operate only on unimodal time series, lack contextual or semantic annotations, and do not support evaluation settings such as cold-start or entity-level generalization. Our work addresses these gaps by introducing a scenario-driven multimodal dataset suite that enables systematic assessment of external modality contributions across diverse forecasting scenarios.

\subsection{Implementation Details} \label{appendix:implement}
In this section, we add details about computation and the prompt.
\subsubsection{Computation Details}
We provide additional details on the time costs associated with our experiments to support reproducibility.
Training time varied significantly across models and datasets, depending on the number of time series, sequence length, and model architecture. Approximate training durations are as follows: Lightweight models such as DLinear and Zero baselines completed training within 1–3 hours. Transformer-based models, like PatchTST, WPMixer, and their multimodal variants, required 6–48 hours per dataset, with longer times on datasets containing long sequences or image/text modalities.
For experiments using GPT-4o-mini, we utilized the OpenAI API. Inference was executed via batch queries with estimated compute cost measured indirectly through input/output token lengths. API calls were rate-limited and parallelized to meet time constraints. We exclude failed, exploratory, or tuning experiments that did not contribute to the final reported results. All reported runs were conducted with fixed random seeds and standardized configurations. 

\subsubsection{Prompt}
The prompt used in cold-start forecasting is shown in Table~\ref{tab:coldstart_prompt}.
\begin{table}[ht]
    \centering
    \small
    \caption{Prompt template used for cold-start forecasting with an LLM. Placeholders, e.g., \texttt{\{horizon\}}, are dynamically filled per instance.}
    \resizebox{0.75\textwidth}{!}{
    \begin{tabular}{p{0.95\linewidth}}
    \toprule
    \textbf{Cold-Start Forecast Prompt} \\
    \midrule
    \textbf{System Prompt (Head)} \\
    \texttt{You are an expert data analyst specializing in \{data\_area\} forecasting. Your task is to generate accurate \{horizon\}-day forecasts for cold-start \{data\_type\} using \{gragality\} \{data\_area\} of relevant \{data\_type\} and text of target \{data\_type\}.} \\
    \hline
    \textbf{User Prompt (Body)} \\
    \texttt{Forecast \{data\_area\} for \{horizon\} days from \{pred\_start\_date\} to \{pred\_end\_date\} beginning from positive value (may include 0 afterwards).} \\
    \texttt{-- Output Format} \\
    \texttt{FORECASTED\_VALUE: [v1, v2, ... v\{horizon\}]} \\
    \texttt{REASONING: A brief explanation (no more than 50 words) referencing the Forecasting Guidelines factors.} \\
    \texttt{-- Target \{data\_type\}} \\
    \texttt{\{target\_text\}} \\
    \texttt{-- Relevant \{data\_type\}} \\
    \texttt{IDs (most similar by \{similar\_method\}): \{relevant\_ids\}} \\
    \texttt{Description: \{relevant\_text\}} \\
    \texttt{\{gragality\} \{data\_area\} from \{relevant\_start\_date\} to \{relevant\_end\_date\}: \{relevant\_series\}} \\
    \texttt{-- Output Example} \\
    \texttt{FORECASTED\_VALUE: [10, 11, 8, 14, 0, 1, 0]} \\
    \texttt{REASONING: xxx} \\
    \texttt{-- Forecasting Guidelines} \\
    \texttt{1. Overall trends of relevant \{data\_area\} before \{pred\_start\_date\}} \\
    \texttt{2. Seasonality patterns (e.g., daily, weekly)} \\
    \texttt{3. Sudden changes or recent shifts} \\
    \texttt{4. External factors affecting this topic (e.g., events, holidays)} \\
    \texttt{5. Historical anomalies} \\
    \texttt{6. Information inferred from related \{data\_type\}s} \\
    \bottomrule
    \end{tabular}}
    \label{tab:coldstart_prompt}
    \end{table}

\subsection{Results}
We present more results in varying-history forecasting and more analysis of the cold-start forecasting performance.

\subsubsection{More Results in Varying-history Forecasting}
\label{sec:appresults}
The results \texttt{mixture} both \texttt{long} and \texttt{short} series, and \texttt{short} series only, are shown in Table~\ref{tab:all_results_2} and Table~\ref{tab:short_results_2}.

\begin{table}[htbp]
    \centering
    \caption{Evaluation on varying-training length forecasting (\texttt{mixture} data with both \texttt{long} and \texttt{short} time series). Best and second-best results are marked based on unrounded values.}
    \label{tab:all_results_2}
    \resizebox{0.8\textwidth}{!}{%
    \begin{tabular}{l|r|rr|rr|rr|rr|rr}
        \hline
        \multicolumn{2}{c|}{Models} & \multicolumn{2}{c|}{DLinear} & \multicolumn{2}{c|}{PatchTST} & \multicolumn{2}{c|}{WPMixer} & \multicolumn{2}{c|}{MultiPatchTST} & \multicolumn{2}{c}{MultiWPMixer} \\\hline
        \multicolumn{2}{c|}{Metric} & rmse & wrmspe & rmse & wrmspe & rmse & wrmspe & rmse & wrmspe & rmse & wrmspe \\
        \toprule
        \multirow{5}{*}{\rotatebox{90}{PixelRec}} & 1 & 1.383 & 1.611 & \underline{1.321} & \underline{1.538} & 1.359 & 1.583 & 1.326 & 1.545 & \textbf{1.314} & \textbf{1.531} \\
        & 7 & 1.566 & 1.821 & 1.462 & 1.700 & 1.476 & 1.716 & \textbf{1.445} & \textbf{1.680} & \underline{1.457} & \underline{1.694} \\
        & 14 & 1.603 & 1.856 & 1.525 & 1.766 & 1.534 & 1.777 & \textbf{1.508} & \textbf{1.747} & \underline{1.519} & \underline{1.760} \\
        & 21 & 1.674 & 1.930 & 1.570 & 1.811 & 1.577 & 1.819 & \textbf{1.554} & \textbf{1.792} & \underline{1.565} & \underline{1.805} \\
        & 28 & 1.676 & 1.925 & 1.609 & 1.848 & 1.621 & 1.861 & \textbf{1.595} & \textbf{1.832} & \underline{1.603} & \underline{1.841} \\
        \hline
        \multirow{5}{*}{\rotatebox{90}{Taobao}} & 1 & 6.969 & 2.011 & \underline{6.641} & \underline{1.916} & 7.194 & 2.075 & 6.952 & 2.006 & \textbf{6.360} & \textbf{1.835} \\
        & 7 & 7.019 & 2.022 & \underline{6.851} & \underline{1.974} & \textbf{6.839} & \textbf{1.971} & 7.114 & 2.050 & 6.937 & 1.999 \\
        & 14 & 7.927 & 2.274 & \textbf{7.063} & \textbf{2.026} & 7.285 & 2.090 & 7.182 & 2.060 & \underline{7.149} & \underline{2.051} \\
        & 21 & 7.822 & 2.245 & 7.324 & 2.103 & 7.444 & 2.137 & \textbf{7.308} & \textbf{2.098} & \underline{7.288} & \underline{2.092} \\
        & 28 & 7.976 & 2.295 & 7.507 & 2.160 & 7.621 & 2.193 & \textbf{7.475} & \textbf{2.151} & \underline{7.490} & \underline{2.155} \\
        \hline
        \multirow{3}{*}{\rotatebox{90}{Tianchi}} & 1 & 202.6 & 12.06 & 199.8 & 11.89 & 201.5 & 11.99 & \underline{196.8} & \underline{11.72} & \textbf{196.7} & \textbf{11.71} \\
        & 7 & 202.4 & 12.31 & 194.7 & 11.84 & 195.9 & 11.92 & \textbf{193.8} & \textbf{11.79} & \underline{193.8} & \underline{11.79} \\
        & 14 & 213.8 & 13.25 & \underline{198.3} & \underline{12.30} & 198.6 & 12.32 & \textbf{198.1} & \textbf{12.28} & 198.4 & 12.30 \\
        \hline
        \multirow{5}{*}{\rotatebox{90}{Amazon}} & 1 & \textbf{0.322} & \textbf{4.036} & 0.323 & 4.052 & 0.326 & 4.078 & \underline{0.323} & \underline{4.045} & 0.325 & 4.071 \\
        & 7 & 0.338 & 4.227 & 0.335 & 4.183 & 0.335 & 4.183 & \textbf{0.334} & \textbf{4.173} & \underline{0.335} & \underline{4.182} \\
        & 14 & 0.345 & 4.290 & 0.343 & 4.264 & \underline{0.342} & \underline{4.256} & \textbf{0.342} & \textbf{4.253} & 0.342 & 4.261 \\
        & 21 & 0.351 & 4.348 & 0.348 & 4.316 & \underline{0.348} & \underline{4.312} & \textbf{0.347} & \textbf{4.307} & 0.348 & 4.317 \\
        & 28 & 0.355 & 4.379 & 0.353 & 4.356 & \underline{0.352} & \underline{4.350} & \textbf{0.352} & \textbf{4.348} & 0.353 & 4.357 \\
        \hline
        \multirow{5}{*}{\rotatebox{90}{Movielens}} & 1 & \textbf{0.230} & \textbf{6.186} & 0.232 & 6.228 & \underline{0.231} & \underline{6.189} & 0.232 & 6.223 & 0.231 & 6.200 \\
        & 7 & \textbf{0.234} & \textbf{6.275} & 0.237 & 6.341 & 0.235 & 6.305 & 0.240 & 6.418 & \underline{0.234} & \underline{6.279} \\
        & 14 & \textbf{0.236} & \textbf{6.309} & 0.241 & 6.436 & \underline{0.237} & \underline{6.348} & 0.244 & 6.513 & 0.238 & 6.369 \\
        & 21 & \textbf{0.237} & \textbf{6.338} & 0.245 & 6.546 & \underline{0.240} & \underline{6.404} & 0.249 & 6.647 & 0.241 & 6.428 \\
        & 28 & \textbf{0.239} & \textbf{6.368} & 0.248 & 6.603 & \underline{0.240} & \underline{6.399} & 0.252 & 6.715 & 0.243 & 6.466 \\
        \hline
        \multirow{5}{*}{\rotatebox{90}{News}} & 1 & 43.71 & 48.11 & \underline{43.24} & \underline{47.59} & 44.56 & 49.04 & \textbf{43.23} & \textbf{47.58} & 44.05 & 48.48 \\
        & 3 & 45.90 & 48.48 & 45.68 & 48.24 & \textbf{45.09} & \textbf{47.63} & 45.47 & 48.02 & \underline{45.13} & \underline{47.67} \\
        & 6 & 39.55 & 41.98 & 39.61 & 42.04 & \underline{39.07} & \underline{41.47} & 39.23 & 41.63 & \textbf{39.06} & \textbf{41.45} \\
        & 9 & 36.02 & 37.85 & 35.90 & 37.72 & \underline{35.47} & \underline{37.26} & 35.75 & 37.57 & \textbf{35.42} & \textbf{37.21} \\
        & 12 & 34.94 & 35.44 & 34.57 & 35.06 & \underline{34.18} & \underline{34.67} & 34.36 & 34.86 & \textbf{34.11} & \textbf{34.60} \\
        \hline
        \multirow{5}{*}{\rotatebox{90}{WikiPeople}} & 1 & 17433 & 4.575 & \underline{16432} & \underline{4.313} & 16456 & 4.319 & 16496 & 4.330 & \textbf{16354} & \textbf{4.292} \\
        & 7 & 17785 & 4.733 & \textbf{17164} & \textbf{4.568} & 17229 & 4.585 & \underline{17170} & \underline{4.569} & 17180 & 4.572 \\
        & 14 & 16421 & 4.401 & \underline{15248} & \underline{4.087} & 15277 & 4.095 & \textbf{15238} & \textbf{4.084} & 15280 & 4.095 \\
        & 21 & 15485 & 4.140 & \underline{14622} & \underline{3.909} & 14691 & 3.927 & \textbf{14609} & \textbf{3.905} & 14653 & 3.917 \\
        & 28 & 15856 & 4.209 & \underline{14506} & \underline{3.851} & 14580 & 3.870 & \textbf{14480} & \textbf{3.844} & 14564 & 3.866 \\
        \bottomrule
    \end{tabular}
    }
\end{table}

\begin{table}[htbp]
    \centering
    \caption{Evaluation on varying-training length forecasting (\texttt{short} time series). Due to rounding to decimal places for table readability, some scores may appear equal but differ in full precision. Best and second-best results are marked based on unrounded values.}
    \label{tab:short_results_2}
    \resizebox{0.8\textwidth}{!}{
    \begin{tabular}{l|r|rr|rr|rr|rr|rr}
        \hline
        \multicolumn{2}{c|}{Models} & \multicolumn{2}{c|}{DLinear} & \multicolumn{2}{c|}{PatchTST} & \multicolumn{2}{c|}{WPMixer} & \multicolumn{2}{c|}{MultiPatchTST} & \multicolumn{2}{c}{MultiWPMixer} \\\hline
        \multicolumn{2}{c|}{Metric} & rmse & wrmspe & rmse & wrmspe & rmse & wrmspe & rmse & wrmspe & rmse & wrmspe \\
        \toprule
        \multirow{5}{*}{\rotatebox{90}{PixelRec}} & 1 & 1.387 & 1.616 & \textbf{1.319} & \textbf{1.536} & 1.361 & 1.585 & 1.324 & 1.542 & \underline{1.314} & \underline{1.530} \\
        & 7 & 1.585 & 1.844 & 1.480 & 1.721 & 1.495 & 1.739 & \textbf{1.464} & \textbf{1.703} & \underline{1.474} & \underline{1.715} \\
        & 14 & 1.628 & 1.887 & 1.548 & 1.795 & 1.559 & 1.808 & \textbf{1.532} & \textbf{1.777} & \underline{1.543} & \underline{1.789} \\
        & 21 & 1.700 & 1.963 & 1.594 & 1.840 & 1.602 & 1.850 & \textbf{1.579} & \textbf{1.823} & \underline{1.589} & \underline{1.835} \\
        & 28 & 1.703 & 1.959 & 1.633 & 1.878 & 1.648 & 1.895 & \textbf{1.619} & \textbf{1.862} & \underline{1.629} & \underline{1.873} \\
        \hline
        \multirow{5}{*}{\rotatebox{90}{Amazon}} & 1 & \textbf{0.263} & \textbf{4.659} & 0.263 & 4.672 & 0.264 & 4.687 &\underline{0.263} & \underline{4.667} & 0.264 & 4.682 \\
        & 7 & 0.273 & 4.821 & 0.269 & 4.754 & 0.269 & 4.751 & \underline{0.269} & \underline{4.751} & \textbf{0.269} & \textbf{4.750} \\
        & 14 & 0.275 & 4.840 & 0.273 & 4.804 & \textbf{0.272} & \textbf{4.797} & 0.272 & 4.801 & \underline{0.272} & \underline{4.800} \\
        & 21 & 0.278 & 4.875 & \underline{0.275} & 4.832 & \textbf{0.275} & \textbf{4.828} & \underline{0.275} & \underline{4.830} & 0.275 & 4.834 \\
        & 28 & 0.279 & 4.883 & 0.277 & 4.852 & \textbf{0.277} & \textbf{4.847} & \underline{0.277} & \underline{4.852} & 0.277 & 4.855 \\
        \hline
        \multirow{5}{*}{\rotatebox{90}{Taobao}} & 1 & 7.659 & 2.188 & \underline{7.336} & \underline{2.095} & 7.897 & 2.255 & 7.650 & 2.185 & \textbf{7.012} & \textbf{2.003} \\
        & 7 & 7.729 & 2.203 & \underline{7.549} & \underline{2.152} & \textbf{7.533} & \textbf{2.148} & 7.834 & 2.233 & 7.639 & 2.178 \\
        & 14 & 8.703 & 2.472 & \textbf{7.750} & \textbf{2.201} & 7.997 & 2.271 & 7.883 & 2.239 & \underline{7.847} & \underline{2.229} \\
        & 21 & 8.476 & 2.413 & 7.969 & 2.269 & 8.078 & 2.300 & \underline{7.943} & \underline{2.261} & \textbf{7.923} & \textbf{2.255} \\
        & 28 & 8.607 & 2.459 & 8.086 & 2.310 & 8.217 & 2.347 & \textbf{8.060} & \textbf{2.303} & \underline{8.076} & \underline{2.307} \\
        \hline
        \multirow{3}{*}{\rotatebox{90}{Tianchi}} & 1 & 202.6 & 12.06 & \underline{199.8} & \underline{11.89} & 201.5 & 11.99 & \textbf{196.8} & \textbf{11.72} & 212.2 & 12.27 \\
        & 7 & 202.4 & 12.31 & \underline{194.7} & \underline{11.84} & 195.9 & 11.92 & \textbf{193.8} & \textbf{11.79} & 211.5 & 12.46 \\
        & 14 & 213.8 & 13.25 & \underline{198.3} & \underline{12.30} & 198.6 & 12.32 & \textbf{198.1} & \textbf{12.28} & 218.5 & 13.06 \\
        \hline
        \multirow{5}{*}{\rotatebox{90}{Movielens}} & 1 & \textbf{0.197} & \textbf{6.259} & 0.198 & 6.295 & \underline{0.197} & \underline{6.265} & 0.197 & 6.273 & 0.197 & 6.274 \\
        & 7 & \underline{0.199} & \underline{6.309} & 0.201 & 6.362 & 0.200 & 6.336 & 0.200 & 6.347 & \textbf{0.198} & \textbf{6.278} \\
        & 14 & \underline{0.200} & \underline{6.321} & 0.203 & 6.415 & 0.201 & 6.348 & 0.203 & 6.397 & \textbf{0.200} & \textbf{6.319} \\
        & 21 & \textbf{0.201} & \textbf{6.332} & 0.206 & 6.481 & 0.203 & 6.383 & 0.206 & 6.476 & \underline{0.202} & \underline{6.347} \\
        & 28 & \textbf{0.203} & \textbf{6.340} & 0.208 & 6.507 & \underline{0.203} & \underline{6.366} & 0.208 & 6.501 & 0.204 & 6.373 \\
        \hline
        \multirow{5}{*}{\rotatebox{90}{News}} & 1 & 22.98 & 37.22 & \underline{21.30} & \underline{34.50} & 23.84 & 38.60 & \textbf{21.27} & \textbf{34.45} & 23.12 & 37.44 \\
        & 3 & 24.81 & 39.45 & 23.95 & 38.09 & \underline{23.39} & \underline{37.20} & 23.59 & 37.51 & \textbf{23.26} & \textbf{36.99} \\
        & 6 & 22.12 & 35.32 & 22.09 & 35.28 & \underline{21.21} & \underline{33.87} & 21.46 & 34.26 & \textbf{21.17} & \textbf{33.80} \\
        & 9 & 20.75 & 33.15 & 20.38 & 32.56 & \underline{19.67} & \underline{31.43} & 20.13 & 32.17 & \textbf{19.56} & \textbf{31.25} \\
        & 12 & 20.78 & 32.86 & 20.02 & 31.65 & \underline{19.29} & \underline{30.51} & 19.64 & 31.06 & \textbf{19.15} & \textbf{30.28} \\
        \hline
        \multirow{5}{*}{\rotatebox{90}{WikiPeople}} & 1 & 11173 & 3.371 & \textbf{10431} & \textbf{3.146} & 10444 & 3.151 & 10479 & 3.161 & \underline{10416} & \underline{3.142} \\
        & 7 & 11872 & 3.620 & 11132 & 3.394 & 11210 & 3.418 & \textbf{11127} & \textbf{3.393} & \underline{11130} & \underline{3.394} \\
        & 14 & 12039 & 3.677 & \underline{11037} & \underline{3.371} & 11056 & 3.377 & \textbf{11021} & \textbf{3.366} & 11055 & 3.377 \\
        & 21 & 11424 & 3.489 & \underline{10750} & \underline{3.284} & 10792 & 3.297 & \textbf{10745} & \textbf{3.282} & 10778 & 3.292 \\
        & 28 & 11809 & 3.592 & \underline{10717} & \underline{3.260} & 10762 & 3.274 & \textbf{10701} & \textbf{3.255} & 10761 & 3.273 \\
        \bottomrule
    \end{tabular}
    }
\end{table}

\subsubsection{Detailed Result Analysis in Cold-start Forecasting}
\label{appendix:coldstart}

This appendix provides dataset-specific analyses for the cold-start forecasting results summarized in Figure~\ref{fig:cold_start}. Each dataset was evaluated using both RMSE and WRMSPE across multiple horizons. Forecasts are generated by GPT-4o-mini conditioned on top-4 retrieved series (based on semantic similarity) and textual metadata. A simple averaging baseline is used for comparison.

For \texttt{AmazonReview}, the performance of \texttt{GPT} and the \texttt{average} baseline is nearly identical across all horizons. This suggests limited benefit from text-based retrieval, potentially due to low inter-series similarity or noisy, generic product descriptions.
For \texttt{Movielens}, \texttt{GPT} shows significantly better performance than \texttt{average} baseline, especially on short horizons. This may be attributed to sparse and spike-like interaction series that lack temporal smoothness. Since the model forecasts each step independently, i.e., without autoregressive dependence, it can exploit retrieved context more flexibly than a naive average.
For \texttt{News}, both methods perform comparably at shorter horizons, with \texttt{GPT} showing slight advantages as the horizon increases. This dataset exhibits strong alignment between textual content and temporal dynamics, e.g., articles on similar topics trending simultaneously, making simple averaging competitive.

For \texttt{TaobaoFashion}, \texttt{GPT} improves WRMSPE in most settings, while RMSE differences remain small. This indicates that the model avoids large relative errors but may still produce some large-scale deviations, possibly due to high demand volatility in short-term behavior.
For \texttt{Tianchi}, a clear performance gap is observed with \texttt{GPT} forecasting. This dataset is both large and sparse, and item-level series often contain zero or near-zero values punctuated by sudden bursts. The model benefits from retrieved examples and contextual descriptions to anticipate such irregularities.
Similar to \texttt{Tianchi}, \texttt{PixelRec} contains long but sparse series. \texttt{GPT} forecasting consistently outperforms averaging in both RMSE and WRMSPE, confirming that external modality information can effectively substitute for missing history when forecasting sparse user-item dynamics.

Overall, the above results confirm that \texttt{MoTime} enables robust evaluation of cold-start forecasting across diverse domains, supporting both retrieval-based and generation-based approaches. Dataset-level behaviors vary depending on sparsity, text quality, and inter-entity similarity, making this benchmark suitable for studying conditional modality utility under extreme data limitations.

\bibliography{ref}
\bibliographystyle{plain}

\end{document}